\documentclass[english]{article}
\usepackage[utf8]{inputenc}

\usepackage{array}
\usepackage{textcomp}
\usepackage{url}
\usepackage{multirow}
\usepackage{amsmath}
\usepackage{amssymb}
\usepackage{graphicx}

\makeatletter

\newcommand{\lyxmathsym}[1]{\ifmmode\begingroup\def\b@ld{bold}
  \text{\ifx\math@version\b@ld\bfseries\fi#1}\endgroup\else#1\fi}

\DeclareTextSymbolDefault{\textquotedbl}{T1}
\providecommand{\tabularnewline}{\\}

\usepackage{hyperref}
\usepackage[linesnumbered]{algorithm2e}
\usepackage[preprint]{neurips_2023}
\renewcommand\cite{\citep}

\makeatother

\usepackage{babel}
\begin{document}
\title{Enhancing Length Extrapolation in Sequential Models with Pointer-Augmented
Neural Memory}
\author{\textbf{Hung Le, Dung Nguyen, Kien Do, Svetha Venkatesh, Truyen Tran
}\\
Applied AI Institute, Deakin University, Geelong, Australia\\
\texttt{thai.le@deakin.edu.au}}
\maketitle
\begin{abstract}
We propose Pointer-Augmented Neural Memory (PANM) to help neural networks
understand and apply symbol processing to new, longer sequences of
data. PANM integrates an external neural memory that uses novel physical
addresses and pointer manipulation techniques to mimic human and computer
symbol processing abilities. PANM facilitates pointer assignment,
dereference, and arithmetic by explicitly using physical pointers
to access memory content. Remarkably, it can learn to perform these
operations through end-to-end training on sequence data, powering
various sequential models. Our experiments demonstrate PANM's exceptional
length extrapolating capabilities and improved performance in tasks
that require symbol processing, such as algorithmic reasoning and
Dyck language recognition. PANM helps Transformer achieve up to 100\%
generalization accuracy in compositional learning tasks and significantly
better results in mathematical reasoning, question answering and machine
translation tasks.
\end{abstract}

\section{Introduction\label{sec:Introduction}}

Systematic generalization underpins intelligence, and it relies on
the ability to recognize abstract rules, extrapolating them to novel
contexts that are distinct yet semantically similar to the seen data.
Current neural networks or statistical machine learning fall short
of handling novel data generated by symbolic rules even though they
have achieved state-of-the-art results in various domains. Some approaches
can show decent generalization for single or set input data \cite{bahdanau2018systematic,gao2020systematic,webb2020emergent}.
Yet, neural networks in general still fail in sequential symbol processing
tasks, even with slight novelty during inference \cite{lake2018generalization,deletang2022neural}.
For instance, these models can easily learn to duplicate sequences
of 10 items, but they will fail to copy sequences of 20 items if they
were not part of the training data. These models overfit the training
data and perform poorly on out-of-distribution samples such as sequences
of greater length or sequences with novel compositions. The issue
also affects big models like Large Language Models, making them struggle
with symbolic manipulation tasks \cite{DBLP:conf/acl/QianWLLY23}.
\emph{This indicates that current methods lack a principled mechanism
for systematic generalization.}

From a neuroscience perspective, it has been suggested that the brain
can execute symbol processing through variable binding and neural
pointers, wherein the sensory data are conceptualized into symbols
that can be assigned arbitrary values \cite{Kriete2013IndirectionAS}.
Like the brain, computer programs excel at symbolic computations.
Programmers use address pointers to dynamically access data or programs,
and have flexible control over the variable. Their programs can work
appropriately with unseen inputs. 

\begin{figure*}
\begin{centering}
\includegraphics[width=0.75\textwidth]{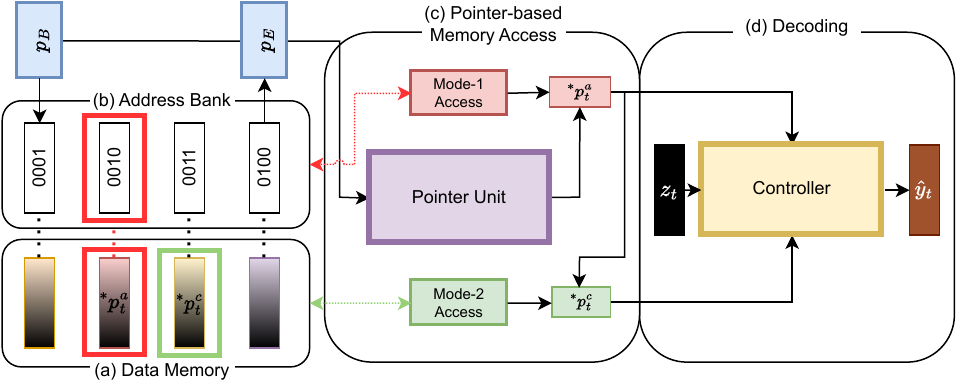}
\par\end{centering}
\caption{PANM architecture. (a) The data memory contains the encoded input
sequence (b) The address bank contains physical addresses associated
with data memory slots. The base and end addresses ($p_{B},p_{E}$)
define the address range of the input sequence. (c) The Pointer Unit
takes $p_{B},p_{E}$, recurrently generates the current pointer $p_{t}^{a}$
and gets its value $^{*}p_{t}^{a}$ via Mode-1 (red)/2 (green) Access.
(d) The Controller takes pointer information, decoding input ($z_{t}=y_{t-}$),
and produce the $t$-th output token $\hat{y_{t}}$. \label{fig:Model-architecture.}}
\end{figure*}
Building on these insights, we propose a pointer-based mechanism to
enhance generalization to unseen length in sequence prediction, which
is a crucial problem that unifies all computable problems \cite{solomonoff2010algorithmic}.\emph{
Our mechanism is based on two principles: }\textbf{(I)} explicitly
modeling pointers as physical addresses, and \textbf{(II)} strictly
isolating pointer manipulation from input data. As such, we need to
design a memory that supports physical pointers, and create a model
that manipulates the pointers to perform abstract rules and access
to the memory. Our memory, dubbed Pointer-Augmented Neural Memory
(PANM), is slot-based RAM \cite{von1993first} where each memory slot
consists of two components: data and address. Unlike initial endeavors
that implicitly model pointers as attention softmax \cite{vinyals2015pointer,kurach2015neural,le2018dual,khan2021deepprocess},
our addresses are generated to explicitly simulate physical memory
addresses, i.e., incremental binary numbers, which is critical for
generalization to longer sequences. 

To manipulate a pointer, we create an address bank that contains physical
addresses corresponding to the input sequence, and use a neural network
called Pointer Unit that is responsible for transforming pointers
from an initial address in the address bank. Through attention to
the address bank, a new pointer is generated as a mixture of the physical
addresses, which can point to different memory slots to follow the
logic of the task. We aim to let the Pointer Unit learn the symbolic
rules of the task in an end-to-end manner. Finally, given a (manipulated)
pointer, the model can access the data through 2 modes of pointer-based
access: pointer dereference (Mode-1) and relational access (Mode-2).
Our memory can be plugged into common encoder-decoder backbones such
as LSTM or Transformer.

Our contribution is a novel memory architecture that incorporates
explicit pointer and symbol processing, working seamlessly with sequential
models to generalize better. We examine our model in symbol-processing
domains such as algorithms and context-free grammar where PANM effectively
works with LSTM and StackRNN. We apply PANM to improve the generalization
of Transformer models on compositional learning, using SCAN and mathematics
datasets. Also, we observe PANM's superior performance in more realistic
question answering and machine translation tasks. Our focus is not
on striving for state-of-the-art results requiring specialized designs
tailored to specific tasks. Our objective is to highlight the generalization
improvement achieved by integrating our memory module into fundamental
sequential models, with minimal architectural changes, and\emph{ showcase
the importance of using fundamental generalizing principles to address
limitations of current deep learning.}

\section{Methods}

\subsection{Problem Formulation}

In sequence-to-sequence (s2s) problems, each data point is an input
sequence $X_{i}=\left\{ x_{t}^{i}\right\} _{t=1}^{l(X_{i})}$, associated
with a target sequence $Y_{i}=\left\{ y_{t}^{i}\right\} _{t=1}^{l(Y_{i})}$
where $l$ is a function returning the length of the sequence. A model
$\Phi$ takes the input $X_{i}$ and generates an output sequence
$\hat{Y}_{i}=\left\{ \hat{y}_{t}^{i}\right\} _{t=1}^{l(\hat{Y}_{i})}$
where the predicted sequence terminates as the model outputs a special
token $\hat{y}_{t=l(\hat{Y}_{i})}^{i}=\mathtt{<eos>}$. Each predicted
token is sampled from a categorical distribution, parameterized by
$\Phi$ and conditioned on the input sequence and optionally with
previous output tokens: $\hat{y}_{t}^{i}\sim p_{\Phi}(y_{t}|X_{i},y_{t-}^{i})$
where $y_{t-}^{i}$ can be $\left\{ y_{k}^{i}\right\} _{k=1}^{t-1}$
(true outputs) or $\left\{ \hat{y}_{k}^{i}\right\} _{k=1}^{t-1}$
(predicted outputs) or even zero, depending on the setting (training
or inference). We train $\Phi$ by minimizing the following cross-entropy
loss: 
\[
\mathcal{L}=\mathbb{E}_{i}\left[-\sum_{t}\log p_{\Phi}(y_{t}^{i}|X_{i},y_{t-}^{i})\right]
\]
We are interested in the ability to handle inputs of arbitrary length,
so we focus on settings in which the length of testing input sequences
is larger than that of training ones: $\max\,l(X_{i})<\min\,l(X_{j})$
with $X_{i}\in\mathcal{D}_{train}$ and $X_{j}\in\mathcal{D}_{test}$.
In the following sections, when there is no confusion, we will drop
the sample index $i$ or $j$ for ease of reading. We note that autoregression
is a special case of the s2s formulation where the input and output
are from the same domain, and the target sequence is one step ahead
of the input sequence. 

\subsection{Pointer Modeling\label{subsec:Pointer-Modeling}}

Computers are powerful thanks to their ability to manipulate pointers.
These pointers store the address of data in memory. Following $\boldsymbol{C}$
programming language notation, let $p$ denote a pointer associated
with a data $d$ in memory $\mathtt{M}$, then $p$ is also the address
of the memory slot containing $d$, i.e., $p=\&d$. We can access
the data via $p$ as $\mathtt{^{*}}p=d$, which is also known as \emph{pointer
dereference}. We can manipulate the pointer to execute various tasks.
For example, if the task is to copy the list $X$ to a new list $Y$,
repeating this simple pointer-based procedure performs correctly regardless
of the list length and the values in $X$: $\mathit{^{*}\&Y={}^{*}\&X;\&Y=\&Y+1;\&X=\&X+1}$. 

In this paper, we propose a way to model pointers by constructing
a bank of addresses, analogous to the addresses in computer architecture.
The address bank starts with a \emph{base address} $p_{B}$ and increases
to form an arithmetic sequence with the common difference of 1. For
example, if the bank has 3 addresses and $p_{B}=3$, the addresses
are $\mathtt{A}=\left\{ 3,4,5\right\} $. We represent the address
as $b$-bit binary vectors, so we have $\mathtt{A}=\left\{ 0010,0011,0100\right\} $
when $b=4$. The address space is $2^{b}$ $b$-bit binary vectors.
Given a memory $\mathtt{M}$ containing $l(\mathtt{M})$ slots, we
bind each slot to an address in the address bank $\mathtt{A}$ such
that $\mathtt{A}[t]=\mathtt{\&M}[t]$ and $^{*}\mathtt{A}[t]=\mathtt{M}[t]$
($1\leq t\leq l(\text{\ensuremath{\mathtt{M}}})$). We use $p_{E}$
as the \emph{end address} to refer to the final address corresponding
to the last item in $\mathtt{M}$.

The memory $\mathtt{M}$ stores the input sequence, and its size depends
on the input length: $l(\mathtt{M})=l(X)$. To enable generalization,
the address space should cover a wide range of addresses that is greater
than the sequence length range (i.e., $2^{b}>\max\,l(X)$). More importantly,
during training, all possible addresses should be exposed to the model.
Otherwise, any unexposed address will confuse the model when it appears
during inference. As such, during training, we \emph{uniformly sample
the base address $p_{B}$} from the address space to construct the
address bank $\mathtt{A}$ for each sequence memory $\mathtt{M}$.
This ensures any address in the address space will eventually appear
between the base and end addresses. See Appendix \ref{subsec:More-Discussion-on}
for an illustrative example of the base address sampling mechanism
and its complexity. 

Given the address bank, we can perform pointer-based procedures to
achieve generalization. To do so, we need pointer variables $p_{t}$
denoting pointers used by the model at timestep $t$. As for the copy
task, the model outputs correctly by accessing and manipulating the
pointer variables through 3 important pointer operations: $p_{t}=\mathtt{A}[t]$
(\emph{assignment}); $\hat{y}_{t}^{i}={}^{*}p_{t}$ (\emph{dereference});
$p_{t}=p_{t}+1$ (\emph{arithmetic}), which will be described in the
next section.

\subsection{Pointer-Augmented Neural Memory (PANM)}

PANM acts as an external memory module for any neural network to support
it handling sequence data. In such a memory-augmented neural network,
a neural Controller ($\mathtt{Ctrl}$) interacts with the memory ($\mathtt{M}$)
to read/write data and make predictions on the output target. Unlike
traditional neural memory, PANM is equipped with an address bank ($\mathtt{A}$)
and a Pointer Unit ($\mathtt{PU}$) to support pointer operations.
To simplify memory writing operations, PANM transforms the whole input
sequence $X$ to the memory $\mathtt{M}$ in the encoding phase such
that $L=l(\mathtt{M})=l(X)$ using $\mathtt{M}=\mathtt{Encoder}_{\theta}\left(X\right)$
where $X\in\mathbb{\mathbb{R}}^{d_{x}\times L}$, $\mathtt{M}\in\mathbb{\mathbb{R}}^{d_{m}\times L}$
and the $\mathtt{Encoder}$, parameterized by $\theta$, can be any
neural encoder such as LSTM or Transformer. The address bank $\mathtt{A}\in\left\{ 0,1\right\} ^{b\times L}$
is then created and bound to $\mathtt{M}$ as mentioned in the previous
section. During decoding, the encoded information in $\mathtt{M}$
is not changed and the controller focuses only on reading from $\mathtt{M}$
to produce the right output sequence. An overview of PANM decoding
process is given in Fig. \ref{fig:Model-architecture.}. 

\subsubsection{Pointer Unit}

At each timestep $t$ of the decoding process, PANM makes use of pointer
variables $p_{t}^{a}$, which are initialized as a valid address in
the address space and then updated by the Pointer Unit $\mathtt{PU}$.
In particular, the $\mathtt{PU}$, implemented as an GRU \cite{chung2014empirical},
takes an address from $\mathtt{A}$ as its initial inputs, e.g., $p_{0}^{a}=p_{B}$,
and recurrently produces a key $h_{t}^{a}$ that performs \emph{address
attention} to create succeeding pointer variables $p_{t}^{a}$:

\begin{align}
h_{t}^{a} & =\mathtt{GRU}_{\varphi}\left(p_{t-1}^{a},h_{t-1}^{a}\right)\\
w_{t}^{a}\left[n\right] & =\mathrm{softmax}\left(\frac{h_{t}^{a}g_{\varphi}^{a}\left(\mathtt{A}[n]\right)}{\left\Vert h_{t}^{a}\right\Vert \left\Vert g_{\varphi}^{a}\left(\mathtt{A}[n]\right)\right\Vert }\right)\\
p_{t}^{a} & =\mathtt{A}w_{t}^{a}\label{eq:pat}
\end{align}
where $h_{0}^{a}$ is initialized as $0$ and $1\leq n\leq l(X)$,
$\varphi$ denotes the parameter of the $\mathtt{PU}$ and $g^{a}\left(\cdot\right)$
is a feed-forward neural network to transform the address to the same
space as $h_{t}^{a}$. According to $\mathsection$ \ref{sec:Introduction}'s
principle \textbf{I}, $p_{t}^{a}$ is ``softly'' \emph{assigned}
a physical address value in the address bank. Our pointer, $p_{t}^{a}$,
offers several advantages over ``implicit pointers'' made up of
the attention weights ($w_{t}^{a}$), which are commonly utilized
in previous works \cite{vinyals2015pointer,luong2015effective}. First,
$p_{t}^{a}$ is a combination of physical addresses represented by
binary numbers, and therefore its dimension is generally independent
of the sequence length. In contrast, the dimension of $w_{t}^{a}$
varies with the input length. Therefore, arithmetic transformations
on $p_{t}^{a}$ are easier than on $w_{t}^{a}$. Second, longer testing
length poses challenges for traditional attentions to accurately produce
$w_{t}^{a}$ pointing to unseen location. Using ``physical key''
$A$ to compute $w_{t}^{a}$ mitigates this issue by employing random
physical address ranges (see $\mathsection$ \ref{subsec:Pointer-Modeling}).

Following $\mathsection$ \ref{sec:Introduction}'s\textbf{ }principle\textbf{
II}, the $\mathtt{PU}$ \emph{recurrently transforms} the original
$p_{0}^{a}$ to a series of pointers $\left\{ p_{t}^{a}\right\} _{t=1}^{l(\hat{Y})}$
suitable for the current task \emph{without using input data}. This
prevents unseen testing inputs disturb $\mathtt{PU}$'s transformations.
In the copy example, an ideal arithmetic transformation ensure $p_{0}^{a}=p_{B}$
and $p_{t+1}^{a}=p_{t}^{a}+1$, which performs perfectly for any sequence
whose length $\leq2^{b}$. We aim to learn $\mathtt{PU}$ to automatically
discover pointer manipulation rules from the task. As the rules are
learned, generalization is achieved even when the testing sequence
is longer or contains novel items. 

\subsubsection{Pointer-based Addressing Modes\label{subsec:Pointer-based-Addressing-Modes}}

\textbf{Mode 1} In this mode, the content from memory $\mathtt{M}$
is retrieved directly by dereferencing pointers. To \emph{dereference
}pointer $p_{t}^{a}$, we utilize the $\mathtt{A}-\mathtt{M}$ binding
and the address attention weight $w_{t}^{a}$, retrieving the pointer
value associated with $p_{t}^{a}$ as $^{*}p_{t}^{a}=\mathtt{M}w_{t}^{a}$.
Through this dereference operator, we can access to arbitrary data
in the memory $\mathtt{M}$ without relying on the content of the
memory. This property enables robustness in handling new sequence
when the memory content is novel and the process stays the same (e.g.,
copy a never-seen-before sequence). Accessing $\mathtt{M}$ indirectly
via $\mathtt{A}$ allows more memory slots to be added to $\mathtt{M}$
during inference without affecting the processing rule as long as
the $\mathtt{PU}$ can transform the pointer variable correctly. During
training, $\mathtt{PU}$ experiences pointer variables covering the
whole address space because the base address is sampled randomly.
Hence, it is possible for $\mathtt{PU}$ to correctly transform pointers
that point to extended addresses of a growing memory as long as the
pointer manipulation rule does not change. The address attention can
be used for multiple pointer variables. In this case, there would
be multiple pointer units $\left\{ \mathtt{PU}_{h}\right\} _{h=1}^{H^{a}}$
responsible for several $\left\{ p_{t,h}^{a}\right\} _{h=1}^{H^{a}}$
and $\left\{ ^{*}p_{t,h}^{a}\right\} _{h=1}^{H^{a}}$ where $H^{a}$
is the number of attention heads. These pointer values will be used
by the Controller for other memory reading. 

\textbf{Mode 2} This mode uses a more complicated memory access mechanism
to capture relations between pointers in complicated reasoning tasks.
\emph{The accessed content is not the one associated with the current
pointer}, but those whose contents are related to the current pointer's
value. As an example, selection sort algorithm may require comparing
items in a list with the Mode-1 pointer's item to select the greater
one. We simulate that using attention with the query as the current
pointer value:

\begin{align}
q_{t} & =g_{\varphi}^{c}\left(\left\{ ^{*}p_{t,h}^{a}\right\} _{h=1}^{H^{a}}\right);\\
w_{t}^{c}\left[n\right] & =\mathrm{softmax}\left(\frac{q_{t}\mathtt{M}[n]}{\left\Vert q_{t}\right\Vert \left\Vert \mathtt{M}[n]\right\Vert }\right)\\
^{*}p_{t}^{c} & =\mathtt{\mathtt{M}}w_{t}^{c}
\end{align}
Here, the pointer attention takes the concatenated values $\left\{ ^{*}p_{t,h}^{a}\right\} _{h=1}^{H^{a}}$
as input, transforms them to a query $q_{t}$ using a feed-forward
neural network $g^{c}\left(\cdot\right)$, and returns the related
pointer value $^{*}p_{t}^{c}$ through attention mechanisms on $\mathtt{M}$.
Intuitively, the Pointer Unit manipulates the Mode-1 pointer $p_{t}^{a}$
such that it retrieves the desired content pointed by the Mode-2 pointer
$p_{t}^{c}$. We can also have multi-head attention, which results
in $\left\{ ^{*}p_{t,h}^{c}\right\} _{h=1}^{H^{c}}$ where $H^{c}$
is the number of attention heads. 

\begin{figure*}
\begin{centering}
\includegraphics[width=1\linewidth]{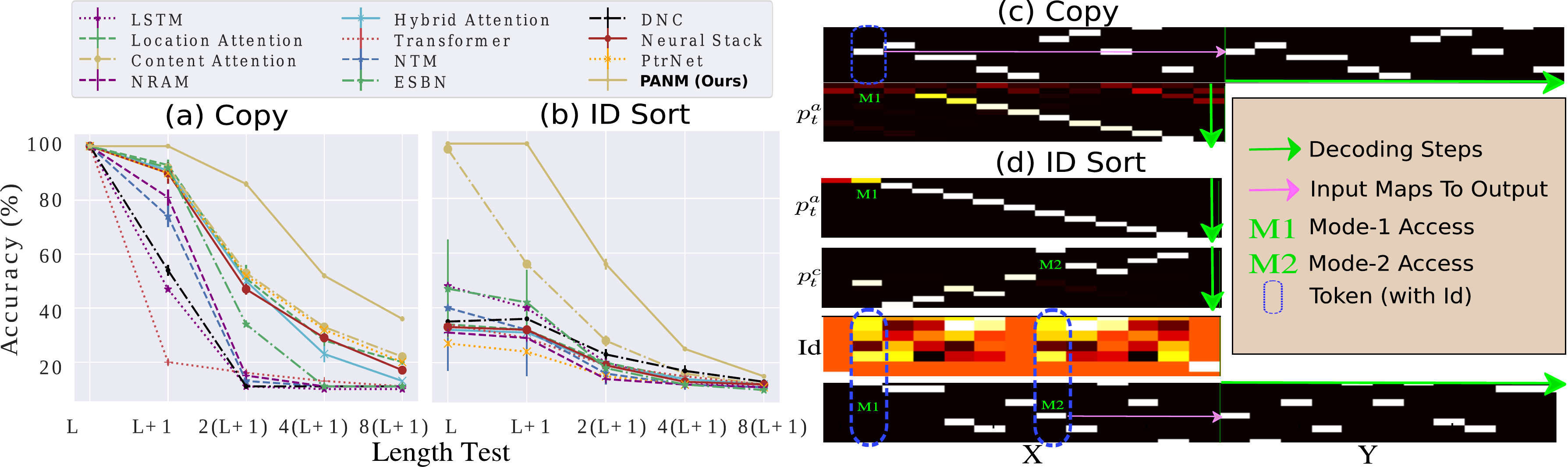}
\par\end{centering}
\caption{Exemplar results on 2 algorithms. (a, b) Test accuracy (mean $\pm$
std) over 5 runs on Copy and ID Sort on each length test, respectively.
Random predictor would reach around 10\% accuracy. (c,d) Visualization
of data and pointer's slots for Copy and ID Sort, respectively. \label{fig:Examplar-performance-on}}
\end{figure*}

\subsubsection{The Controller \label{subsec:The-Controller}}

The Controller $\mathtt{Ctrl}$ is responsible for decoding the memory
to produce outputs. Unlike other methods, we have pointer-based memory
access to provide the controller with symbol-processing information.
In particular, at the $t$-th step of the decoding, $\mathtt{Ctrl}$
takes the pointer values (mode 1 and 2) as input together with an
optional decoding input ($z_{t}=y_{t-}$), and uses a GRU to recurrently
produce the hidden state $h_{t}^{c}$ as follows,

\begin{align}
h_{t}^{c} & =\text{\ensuremath{\mathtt{GRU}}}_{\lambda}\left(\left[\left\{ ^{*}p_{t,h}^{a}\right\} _{h=1}^{H_{a}},\left\{ ^{*}p_{t,h}^{c}\right\} _{h=1}^{H_{c}},z_{t}\right],h_{t-1}^{c}\right)\label{eq:ctr}
\end{align}
where the hidden state $h_{0}^{c}$ is initialized as $\sum_{i}\mathtt{M}[i]$
and $\lambda$ is the parameters of $\mathtt{Ctrl}$. The $\mathtt{GRU}$
handles content-based input, empowered with pointer information to
construct rich hidden state representations. Furthermore, the pointer-based
data gives the $\mathtt{GRU}$ access to correct items in the sequence
even when the memory content becomes different from the training due
to encountering novel sequence length. 

The Controller $\mathtt{Ctrl}$ uses the pointer values (mode 1),
the related pointer values (mode 2) and the hidden state $h_{t}^{c}$
to make the final prediction. It simply concatenates them and forward
to the $g^{o}\left(\cdot\right)$--a MLP, to generate the output
token 
\[
\hat{y}_{t}^{i}\sim p_{\Phi}(y_{t}|X_{i},z_{t})=g_{\lambda}^{o}\left(\left[\left\{ ^{*}p_{t,h}^{a}\right\} _{h=1}^{H^{a}},\left\{ ^{*}p_{t,h}^{c}\right\} _{h=1}^{H^{c}},h_{t}^{c}\right]\right)
\]
The pointer values allow $g^{o}$ to fully utilize pointer information
in producing the final output.  $\Phi$ consists of the parameters
of the $\mathtt{Encoder}_{\theta}$, Pointer Unit $\mathtt{PU}_{\varphi}$
and Controller $\mathtt{Ctrl}_{\lambda}$. $\mathtt{Ctrl}$ can be
put on top of another decoder to process the decoding input $z_{t}$.
In some experiments, we use Transformer as the decoder (see Appendix
\ref{subsec:Choice-of-Architecture} and \ref{subsec:Conpositional-Learning-App}).
A summary of PANM's operation is given in Algo. \ref{algo} in Appendix.

\section{Experimental Results}

In our experiments, we use two pointer variables in Mode-1 access
and one for Mode-2 to balance between performance and computing cost
($H^{a}=2$, $H^{c}=1$, see more in Appendix \ref{subsec:Choice-of-Architecture}).
The two Mode-1 pointer variables are initialized as the base and end
addresses. All MLPs in PANM have 1 hidden layer of 128 dimensions.
We use 256-dimensional GRUs for $\mathtt{PU}$ and $\mathtt{Ctrl}$.
The memory's address space has $b=10$ bits, corresponding to a maximum
of 1024 unique addresses, which is greater than any sequence length
in the experiments. 

In $\mathsection$\ref{subsec:Algorithmic-Reasoning-1}-\ref{subsec:Compositional-Learning},
our chosen tasks are representative of various types of symbolic reasoning
and well-known benchmarks to measure the symbol-processing capability
of ML models. \emph{To showcase that these tasks are non-trivial,
we report how Chat-GPT failed on our tasks in Appendix} \ref{subsec:Failures-of-Chat-GPT}.
We validate the contribution of our methods in other practical tasks
in $\mathsection$\ref{subsec:Other-NLP-Tasks}. We also explain the
choice of competing baselines in Appendix \ref{subsec:Experimental-Details}. 

\begin{table*}
\begin{centering}
\begin{tabular}{ccccccc}
\hline 
Task & Copy & Reverse & Mix & D. Recall & P. Sort & ID Sort\tabularnewline
\hline 
Other Max & 60.2 & 63.6 & 64.0 & 47.6 & 60.6 & 42.1\tabularnewline
\hline 
PANM (Ours) & \textbf{74.8} & \textbf{73.6} & \textbf{81.2} & \textbf{52.8} & \textbf{67.8} & \textbf{59.2}\tabularnewline
\hline 
\end{tabular}
\par\end{centering}
\caption{Algorithmic reasoning: mean sequence-level accuracy (\%) over testing
lengths Other Max is selected as the best numbers at each length mode
from other baselines. \label{tab:Algorithmic-reasoning:-mean}}
\end{table*}
\begin{table*}
\begin{centering}
{\scriptsize{}}%
\begin{tabular}{c|ccccccccccc|cc}
\hline 
\multirow{2}{*}{Task} & \multicolumn{11}{c|}{SCAN (L cut-off)} & \multicolumn{2}{c}{Math}\tabularnewline
\cline{2-14} 
 & 22 & 24 & 25 & 26 & 27 & 28 & 30 & 32 & 33 & 36 & 40 & $\mathtt{a.s}$ & $\mathtt{p.v}$\tabularnewline
\hline 
U. TRM+RPE & 20 & 12 & 71 & \textbf{100} & \textbf{100} & \textbf{100} & \textbf{100} & \textbf{100} & \textbf{100} & \textbf{100} & \textbf{100} & \textbf{97} & 75\tabularnewline
TRM+RPE & 20 & 12 & 31 & 61 & \textbf{100} & \textbf{100} & \textbf{100} & 94 & \textbf{100} & \textbf{100} & \textbf{100} & 91 & 0\tabularnewline
U. TRM & 2 & 5 & 14 & 21 & 26 & 0 & 6 & 35 & 0 & 0 & 0 & 94 & 20\tabularnewline
TRM & 0 & 4 & 19 & 29 & 30 & 8 & 24 & 36 & 0 & 0 & 0 & 89 & 12\tabularnewline
\hline 
PANM (Ours) & \textbf{22} & \textbf{47} & \textbf{100} & \textbf{100} & \textbf{100} & \textbf{100} & \textbf{100} & \textbf{100} & \textbf{100} & \textbf{100} & \textbf{100} & \textbf{97} & \textbf{86}\tabularnewline
\hline 
\end{tabular}{\scriptsize\par}
\par\end{centering}
\caption{SCAN (Left): Exact match accuracy (\%, median of 5 runs) on splits
of various lengths. Mathematics (Right): mean accuracy over 5 runs.
The baselines' numbers are from \citet{csordas2021devil} and we run
PANM using the authors' codebase. \label{tab:SCAN:-Exact-match}\label{tab:Mathematics-dataset:-mean}}
\end{table*}

\subsection{Algorithmic Reasoning\label{subsec:Algorithmic-Reasoning-1}}

In our first experiment, we study the class of symbol processing problems
where an output sequence is generated by a predefined algorithm applied
to any input sequence (e.g., copy and sort). The tokens in the sequences
are symbols from $0$ to $9$. The input tokens can be coupled with
meta information related to the task such as the priority score in
Priority Sort task. During training, the input sequences have length
up to $L$ tokens and can grow to $L+1,$ $2(L+1)$, $4(L+1)$ or
$8(L+1)$ during testing. Our setting is more challenging than previous
generalization tests on algorithmic reasoning because of four reasons:
(1) the task is 10-class classification, harder than binary prediction
in \citet{graves2014neural, le2022neurocoder}, (2) the testing data
can be eight time longer than the training and the training length
is limited to $L\approx10$, which is harder than \citet{grefenstette2015learning},
(3) there is no curriculum learning as in \citet{kurach2015neural},
and (4) the training label is the one-hot value of the token, which
can be confusing in case one token appears multiple times in the input
sequence and tougher than using label as the index/location of the
token as in \citet{vinyals2015pointer}. 

Here, we design several tasks\textbf{. Content-free tasks }involve
permuting tokens in input sequence using certain position-based rules:
First-In-First-Out (\emph{Copy}), Last-In-First-Out (\emph{Reverse})
and \emph{Mix}. While the first two rules demand linear pointer manipulations
(traverse the sequence from the head or tail, to output the target
token), the last one uses a non-linear, length-dependent manipulation
rule: if $t$ is odd, $y_{t}=x_{\left\lceil \frac{L}{2}\right\rceil }$;
if $t$ is even, $y_{t}=x_{1}$. \textbf{Content-based tasks }need
the input's token value together with symbol processing to arrange
the output sequence. We introduce 3 tasks: \emph{Dynamic Recall},
\emph{Priority Sort} and \emph{ID Sort}. Readers can find the details
of these tasks in Appendix \ref{subsec:Algorithmic-Reasoning}.

\textbf{Baselines }are categorized into 4 groups: (1) Traditional
RNNs such as LSTM \cite{hochreiter1997long}, (2) Sequential attention
models: Content Attention \cite{bahdanau2014neural}, Location Attention
\cite{luong2015effective}, Hybrid Attention (our baseline concatenates
the attention vectors from content and location attention), (3) MANNs
such as NTM \cite{graves2014neural}, DNC \cite{graves2016hybrid},
Neural Stack \cite{grefenstette2015learning} and Transformer \cite{vaswani2017attention},
and (4) pointer-aware models: NRAM \cite{kurach2015neural}, PtrNet
\cite{vinyals2015pointer}, ESBN \cite{webb2020emergent} and our
method PANM. In this synthetic experiment, we adopt LSTM as the encoder
for PANM. All baselines are trained with fixed number of steps (100K
for ID Sort and 50K for the rest), which is enough for the training
loss to converge. For each task, each baseline is trained 5 times
with different random seeds and we use the best checkpoint on $L+1$
mode validation to evaluate the baselines. 

\textbf{Results }We report the average accuracy across different testing
length for each task in Table \ref{tab:Algorithmic-reasoning:-mean}.
Overall, PANM significantly outperforms the best competitors ranging
from 10-20\% per task. Compared with individual baselines, the improvement
is much higher (Appendix \ref{subsec:Algorithmic-Reasoning}). We
illustrate how the pointer manipulation works for Copy and ID Sort
in Fig. \ref{fig:Examplar-performance-on} (c) and (d). In Copy, only
Mode-1 access is needed. As decoding step $t$ increases, Pointer
Unit generates $p_{t}^{a}$ following the increment of the addresses
as expected. In ID Sort, both Mode-1 and 2 are needed. The Pointer
Unit generates $p_{t}^{a}$ incrementally to trace the input tokens
from left to right (Mode 1). Then, the Mode-2 pointer $p_{t}^{c}$
is computed via attention to discover token with the same id, which
will be the output at step $t$. Without Mode-2 access, PANM certainly
fails this task. Experiments with varied number of heads are in Appendix
\ref{subsec:More-Experiments-with}. 

\begin{figure*}
\begin{centering}
\includegraphics[width=0.9\linewidth]{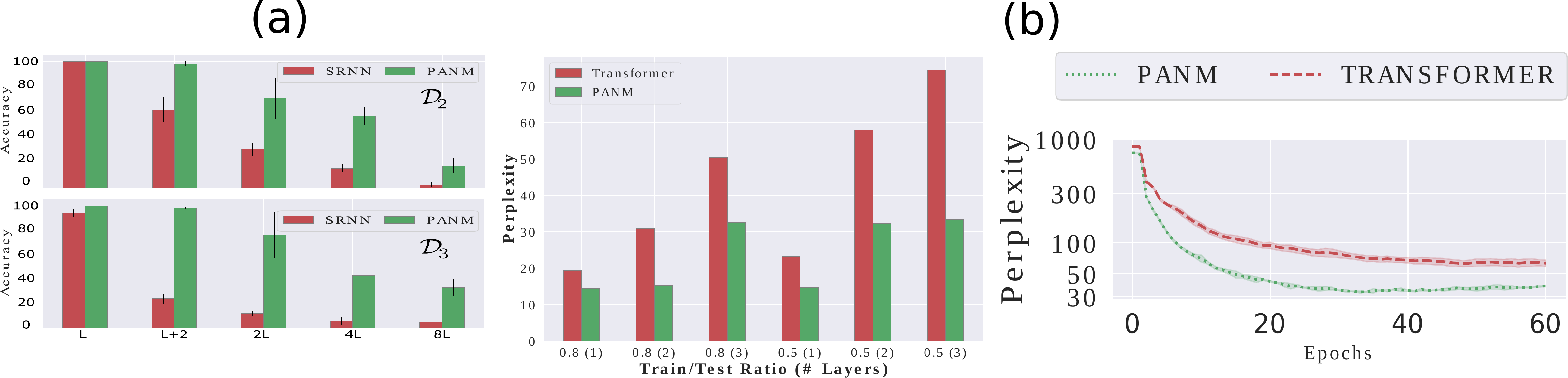}
\par\end{centering}
\caption{(a) Dyck: mean $\pm$ std. accuracy over 5 runs with different testing
lengths. (b) Machine translation task: Perplexity on Multi30K dataset
(the lower the better). We sort the sequences in the data by length
and create 2 settings using train/test split of 0.8 and 0.5, respectively.
The baselines are Transformer and PANM. \textbf{Left:} The best test
perplexity over 2 settings for different number of Transformer's layers
(1 to 3 layers). \textbf{Right:} an example of testing perplexity
curves over training epochs for the case of 0.5 train/test split (2
layers) where we run 3 times and report the mean$\pm$std. The y-axis
is visualized using log scale.\label{fig:Additional-results-on}\label{fig:dyck-1}}
\end{figure*}

\subsection{Dyck Language Recognition\label{subsec:Dyck-Language-Recognition}}

Truly understanding the hidden law of context-free grammars such as
Dyck ($\mathcal{D}_{n}$) is challenging for neural networks, even
those with memory and attention \cite{yu2019learning}. The language
consists of strings with balanced pairs of brackets of different types
($|_{1},|_{2}$,...,$|_{n}$), generated by the following rules: $S\rightarrow|_{i}S|_{i}$
with probability $p/n$ or $SS$ with probability $q$ or $\epsilon$
with probability $1-p-q$. Here, $p,q$ are parameter of the language
and $\epsilon$ is equivalent to EOS token. We follow the sequence
prediction task and datasets in \citet{suzgun2019memory} where the
input is an unfinished Dyck string, and the output is the set of possible
brackets for the next token, e.g., for $\mathcal{D}_{2}$, $\mathtt{(}[]\rightarrow\mathtt{(}$
or $\mathtt{)}$ or $\mathtt{[}$. We follow the authors to enable
set prediction by representing output $y_{t}$ as a multi-hot vector.

We adapt PANM to this autoregression task by masking $\mathtt{M}$
to ensure the decoder only see tokens up to the current decoding step.
Since the token set is simple, we do not need to use any encoder,
i.e., raw input tokens are stored in $\mathtt{M}$. The SOTA baseline
in this task is SRNN \cite{suzgun2019memory}, an autoregressive model
using stack as memory. We use this model as the decoder to make the
setup of PANM close to SRNN. The only difference is that PANM has
Pointer-Based Memory Access (Fig. \ref{fig:Model-architecture.} (b)).
To make the task more challenging, we limit the maximum training length
$L$ to 10 ($\mathcal{D}_{2}$) and 20 ($\mathcal{D}_{3}$) tokens,
and the testing lengths are $L+2$, $2L$, $4L$, $8L$. We choose
$L$ as minimum numbers such that the model can perform decently on
training data. The standard training and testing sizes are 5000. We
train the models for 5 epochs and evaluate them on the training data
at the end of each epoch to save model checkpoints. We use the best
checkpoints for generalization test. 

Fig. \ref{fig:dyck-1} (a) reports the models' accuracy for $\mathcal{D}_{2}$
and $\mathcal{D}_{3}$. Under our extreme setting, SRNN generalization
fades out quickly as test lengths increase, especially for $\mathcal{D}_{3}$
whereas PANM performance degradation happens at a much slower rate,
outperforming SRNN by around 20\% on average in both tasks at any
test lengths.

\subsection{Compositional Learning\label{subsec:Compositional-Learning}}

\textbf{SCAN} In this task , one needs to map an input sentence into
an output sequence of commands \cite{lake2018generalization}. The
sequences are compositional, consisting of reusable parts. For example,
in one case, $\mathtt{\lyxmathsym{\textquotedblleft}jump\,twice\lyxmathsym{\textquotedblright}}$
should be mapped to $\mathtt{\lyxmathsym{\textquotedblleft}JUMP\,JUMP\lyxmathsym{\textquotedblright}}$
and in another, $\mathtt{\lyxmathsym{\textquotedblleft}walk\,twice\lyxmathsym{\textquotedblright}}$
becomes $\mathtt{\lyxmathsym{\textquotedblleft}WALK\,WALK\lyxmathsym{\textquotedblright}}$.
We focus on the ``length split'' datasets where the training sequences
are shorter than the test ones with 11 length modes $L=22,24,..,40$
\cite{newman2020extrapolation}. We adopt the benchmark, training
procedure and baselines prepared by \citet{csordas2021devil}, which
achieves strong results under standard s2s learning. Here, our aim
is not to break SOTA, which can be achieve by hybrid-symbolic architectures
\cite{chen2020compositional,shaw2021compositional}. Instead, we focus
on improving Transformer generalization in this task, hence the baselines
are chosen as several variants of Transformers (TRM) targeted to sequence
extrapolation, including those using Relative Positional Encoding
(RPE \cite{dai2019transformer}) and Universal Transformer (U. TRM
\cite{dehghaniuniversal}), which is an advanced Transformer variant
that recurrently processes each token, and can dynamically adjust
the number of processing steps. Following \citet{csordas2021devil},
each baseline is trained 5 times for 50K steps and the resulting model
after training is used for evaluation (no validation). Here, we use
Transformer as the $\mathtt{Encoder}$, which is the same as the TRM,
and stack the Controller to another Transform decoder (see details
in Appendix \ref{subsec:Conpositional-Learning-App}). Hence, the
only difference is the decoding where PANM leverages pointer manipulation. 

Table \ref{tab:SCAN:-Exact-match} shows that PANM outperforms other
baselines in the hardest settings when the training length is up-to
22, 24, and 25. For 22 and 24 cases, general models like PANM cannot
show perfect generalization because some testing compositions is entirely
discarded from the train set. In easier settings, PANM shares the
perfect median accuracy with the sophisticated U. TRM + RPE although
it does not use RPE. Remarkably, despite sharing the same encoder,
TRM performs much worse than PANM and even fails to learn in easy
modes (33, 36, 40), indicating the importance of pointer handling
in this testbed. One problem for other baselines is the EOS~decision
(when to generate ending token), which requires length tracking \cite{newman2020extrapolation}.
As they do not have content-free sequence iteration mechanisms, it
is extremely hard to trace the length without overfitting to the training
data. On the other hand, PANM can hypothetically generate pointer
incrementally and capture the difference between the last and the
first pointers, i.e. the input length, and infer the output sequence
length based on that information. 

\begin{table}
\begin{centering}
\begin{tabular}{ccc}
\hline 
\multirow{2}{*}{Model} & \multicolumn{2}{c}{Split}\tabularnewline
 & 0.8-0.2 & 0.5-0.5\tabularnewline
\hline 
Transformer & 0.79 \textpm{} 0.01 & 0.76 \textpm{} 0.01\tabularnewline
U. TRM+ RPE & 0.80 \textpm{} 0.02 & 0.75 \textpm{} 0.01\tabularnewline
PANM (Ours) & \textbf{0.85 \textpm{} 0.02} & \textbf{0.81 \textpm{} 0.03}\tabularnewline
\hline 
\end{tabular}
\par\end{centering}
\caption{bAbI QA: Testing accuracy (mean $\pm$ std.) over 5 runs.\label{tab:bAbI-QA:-testing}}
\end{table}
\textbf{Mathematical Problems} We test our model on mathematics \cite{saxton2018analysing}
where the input/output are questions and answers about math and each
token is a character. For example, $\mathtt{What\,\,is-5-110911?\rightarrow-110916}$
(add\_or\_sub) and $\mathtt{What\,\,is\,\,the\,\,hundreds\,\,digit\,\,of\,\,31253?\rightarrow2}$
(place\_value). The task requires not only math reasoning, but also
natural language understanding. We follow the training from \citet{csordas2021devil}
to conduct experiments on 2 subsets: $\mathtt{add\_or\_sub}$ (a.s)
and $\mathtt{place\_value}$ (p.v), and compare our method with Transformer-based
baselines. Here, we focus on the extrapolating test set involving
larger numbers, more numbers, more compositions, and thus longer input
sequences than the training. We use TRM + RPE as the $\mathtt{Encoder}$
and the Controller is added to a normal TRM decoder. As shown in Table
\ref{tab:Mathematics-dataset:-mean}, on $\mathtt{place\_value},$
PANM does not suffer from performance crash as TRM + RPE (0\% test
accuracy, as admitted in the paper \cite{csordas2021devil} even though
it uses the same encoder). PANM achieves similar results as U. TRM+
RPE on $\mathtt{\mathtt{add\_or\_sub}}$ while outperforming it by
11\% on $\mathtt{place\_value}$. We also examine PANM with the original
Transformer and report additional results in Appendix \ref{subsec:Conpositional-Learning-App}.

\begin{table}
\begin{centering}
\begin{tabular}{ccccc}
\hline 
\multirow{3}{*}{Model} & \multicolumn{4}{c}{Split}\tabularnewline
 & \multicolumn{2}{c}{0.8-0.2} & \multicolumn{2}{c}{0.5-0.5}\tabularnewline
 & F1 & EM & F1 & EM\tabularnewline
\hline 
BERT & 0.77 & 0.64 & 0.73 & 0.59\tabularnewline
PANM (Ours) & \textbf{0.78} & \textbf{0.65} & \textbf{0.76} & \textbf{0.61}\tabularnewline
\hline 
\end{tabular}
\par\end{centering}
\caption{SQUAD 1.1: Testing accuracy after 3 epoch fine-tuning. F1 score and
exact match (EM) follows the standard evaluation in \citet{kenton2019bert}.
\label{tab:bAbI-QA:-testing-1}}
\end{table}

\subsection{Other NLP Tasks\label{subsec:Other-NLP-Tasks}}

\textbf{Question Answering }Our objective is to explore the PANM's
generalization beyond obviously compositional data by applying it
in a more practical setting of question answering. For this purpose,
we utilize two datasets, namely bAbI \cite{weston2015towards} and
SQUAD 1.1 \cite{rajpurkar2016squad} where the input sequence is a
context paragraph and a question, and the output is the answer. To
add complexity to the task, we ensure the length of test sequence
is greater than that of the training by sorting the context paragraph
by length and splitting the sorted data into 0.8/0.2 and 0.5/0.5 ratio.
Details of the data/task are in Appendix \ref{subsec:Question-Answering-App}.
In \textbf{bAbI}, we configure the PANM similarly to the one described
in $\mathsection$ \ref{subsec:Compositional-Learning} using Transformer
backbone, and test the models after 100-epoch training. The models
predict the answer tokens given the context and question tokens. As
shown in Table \ref{tab:bAbI-QA:-testing} and Appendix Fig. \ref{fig:dyck}
(right), PANM helps Transformer generalize better, consistently improving
around 6\% and 5\% using 0.8/0.2 and 0.5/0.5 splits, respectively.
Notably, PANM's testing loss is not diverged quickly as Transformer's,
indicating PANM's capability of reducing overfitting. In \textbf{SQUAD},
we use BERT as the backbone to predict the start and the end of the
answer as in \citet{kenton2019bert}. PANM-assisted model outperforms
the baselines by 1\% and 2\% exact match accuracy, respectively (Table
\ref{tab:bAbI-QA:-testing-1}). The improvement is significant as
BERT is a big foundation model already pretrained with big data and
robust against novel test data. 

\textbf{Machine Translation }Here, we want to verify the PANM in machine
translation and show that PANM can work with different number layers
of Transformer. The results are presented in Fig. \ref{fig:dyck-1}
(b) where we report the model perplexity on Multi30K (en-de) dataset.
The 30K-sample dataset is sorted by input length and split into training
and testing s.t. testing sequences are longer, similar to QA task.
The results demonstrate PANM can consistently improve the generalization
performance of  Transformer across different split ratios and the
number of encoder/decoder layers.

\section{Related works}

There are many attempts to augment neural networks with external memory
(MANN) to improve their symbol-processing ability. Pioneers such as
NTM \cite{graves2014neural} and DNC \cite{graves2016hybrid} propose
computer-like memory read/write operations with content-based attention
mechanisms, and thus in principle, can execute any symbolic rules.
However, learning the hidden law end-to-end from sequence data is
extremely difficult. Therefore, MANNs including Transformers \cite{vaswani2017attention},
may fail miserably in out-of-distribution testbeds, especially length
extrapolation \cite{deletang2022neural}. Recent LLMs are good at
reasoning and generalization, but bad at symbolic processing \cite{DBLP:conf/acl/QianWLLY23,tang2023large}.
We use LLMs only to show our task difficulty (Appendix \ref{subsec:Failures-of-Chat-GPT}),
not as a baseline, because they are not on the same scale as our method.

Many recent works advocate the use of specialized memory architectures
such as stacks \cite{grefenstette2015learning,hao2018context,suzgun2019memory},
key-value memory \cite{webb2020emergent,Le2020Neural} and improved
attentions \cite{kurach2015neural,russin2019compositional,dubois2020location}.
These methods employ different inductive biases in designing the memory
and attention, \emph{yet not following the two principles advocated
by our paper}. Although they may work remarkably on certain synthetic
tasks, they are not examined on various benchmarks or compatible with
different sequential backbones. Other orthogonal approaches focus
on model initialization \cite{zhang2019improving}, data augmentation
\cite{andreas2020good} or training details \cite{csordas2021devil}.
Besides differentiable models, there are major progress in compositional
rule learning that leverage neuro-symbolic architectures \cite{nye2020learning,shaw2021compositional,chen2020compositional}
or reinforcement learning \cite{liu2020compositional}. We have not
compared our model with these task-specific methods, as our focus
is on improving the systematic generalization of fundamental differentiable
models.

Our approach is mainly related to key-value memory because the address
bank can be viewed as the key and the data memory as the value. However,
the key in other works is either learned through backpropagation \cite{Le2020Neural}
or computed based on the input data \cite{webb2020emergent}. In contrast,
our ``keys'' are generated as fixed numbers (physical memory addresses--
$\mathsection$ \ref{sec:Introduction}'s principle \textbf{I}), which
is totally separated from the data and extendable to longer sequences.
We argue that using addresses as keys is critical to symbol processing
because it explicitly allows pointer assignment, dereference and arithmetic.
A related generalization-enable scheme is to design positional encoding
of tokens in a sequence \cite{vaswani2017attention,dai2019transformer,li2022systematic}.
Unlike these approaches, our physical addresses are detached from
the data to support transforming pointers through time steps and isolating
pointer manipulation from the input.

\section{Discussion}

We introduce a neural memory model called PANM that manipulates pointers
to learn symbol processing rules for better length extrapolation.
PANM isolates symbols from data and uses an address bank to allow
data-isolated pointer manipulation through address attention. PANM
consistently outperforms strong baselines in tasks such as algorithm
mining, compositional learning, mathematics reasoning, context-free
grammar recognition, and practical NLP tasks, even when the test sequence
is much longer than the training sequence.\textbf{ Reproducibility
}In the Appendix, we included our detailed model descriptions, algorithms,
implementations, hyperparameters to replicate the results of our experiments.
Source code will be released when the paper is published\textbf{. }

\paragraph{Impact Statements }

In this work, we used the publicly available datasets for experiments.
We did not collect human or animal data during this study. Our work
aims to improve the generalization of sequential models. This aim
is genuine, and we do not think there are immediate harmful consequences.
However, we are aware of potential problems if our method is used
to augment language models to generate texts that are hallucinated
or contain negative contents. This issue is typical for plug-and-play
modules like PANM, and we will do our best to prevent it from our
end.

\bibliographystyle{plainnat}
\bibliography{nam}

\cleardoublepage{}

\section*{Appendix}

\renewcommand\thesubsection{\Alph{subsection}}

\subsection{More Discussion on Related Works}

The proposed address attention in our paper is comparable to two known
mechanisms: (1) location-based attention \cite{luong2015effective,dubois2020location}
and (2) memory shifting \cite{graves2014neural,yang2016lie}. The
former uses neural networks to produce attention weights to the memory/sequence,
which cannot help when the memory grows during inference since the
networks never learn to generate weights for the additional slots.
Inspired by Turing Machine, the latter aims to shift the current attention
weight associated with a memory slot to the next or previous slot.
Shifting-like operations can handle any sequence length. However,
it cannot simulate complicated manipulation rules. Unlike our $\mathtt{PU}$
design which obeys $\mathsection$ \ref{sec:Introduction}'s principle
\textbf{II}, the attention weight and the network trained to shift
it depend on the memory content $\mathtt{M}$. That is detrimental
to generalization since new content can disturb both the attention
weight and the shifting network as the memory grows. 

Another line of works tackles systematic generalization through meta-learning
training \cite{lake2019compositional}, while our method employs standard
supervised training. These approaches are complementary, with our
method concentrating on enhancing model architecture rather than training
procedures, making it applicable in diverse settings beyond SCAN tasks.
Additionally, the study by Hu et al. (2020) addresses syntactic generalization
\cite{hu2020systematic}, a different problem compared to our paper,
which emphasizes length extrapolation across various benchmarks. Notably,
our paper considers similar baselines, such as LSTM and Transformer,
as those examined in the referenced papers. There are other lines
of research targeting reasoning and generalization using image input\cite{wu2020analogical,eisermann2021generalization}.
They are outside the scope of our paper, which specifically addresses
generalization for longer sequences of text or discrete inputs

Our address bank and physical pointers can be viewed as some form
of positional encoding. However, we do not use simple projections
or embeddings to force the attention to be position-only. Instead,
we aim to learn a series of transformations that simulate the position-based
symbolic rules. At each time step, a new pointer (\textquotedbl position\textquotedbl )
is dynamically generated that reflects the manipulation rule required
by the task (e.g. move to the next location), which is unlike the
positional encoding approaches such as RPE \cite{dai2019transformer}
which aims to provide the model with information on the relative position
or distance of the timesteps. We summarise the difference between
our method and Transformer in Table \ref{tab:compare_T}. 

\subsection{More Discussion on Base Address Sampling Mechanism\label{subsec:More-Discussion-on}}

We provide a simple example to illustrate how base address sampling
help in generalization. Assume the training sequence length is 10,
and the desired manipulation is $p'=p+1$ (copy task). Assume the
possible address range is ${0,1,...,19}$, which is bigger than any
sequence length. If $p_{B}=0$ , the training address bank contains
addresses: ${0,1,...8,9}$. Without base address sampling, the model
always sees the training address bank of ${0,1,...8,9}$ and thus
can only learn manipulating function for $0\leq p\leq9$, thereby
failing when testing address bank includes addresses larger than 9. 

Thanks to base address sampling, at some point of training, $p_{B}=10$
, the training address bank is ${10,11,...13,19}$. The manipulating
function sees $p>9$ and can learn to transform $p'=p+1$ for $p>9$,
e.g., transform $p=10\rightarrow p'=11$. The learning happens because
the pointer's value ($^{*}p$) is used to predict the output sequence.
The task loss will reward $^{*}p$ that follows the rule, and update
the Pointer Unit such that it transforms the $p$ following the rule
to minimize the loss. During testing, the input length can be 12,
we set $p_{B}=0$ and the address bank is ${0,1,....,10,11}$. The
learned manipulation can still be applied to new locations 10th, and
11th.

We can prove that the complexity of exposing all addresses to the
model is practically small compared to the normal training. Assume
the training input sequence length is $L$, and the number of possible
addresses is $L_{max}$. Here, $L_{max}$ indicates the possible testing
length that the model can handle. When $L_{max}\rightarrow\infty$
, the expected number of samples required for exposing all addresses
is $O\left(n\log n\right)$ where $n=L_{max}/L$ (we can formulate
this problem as Coupon collector's problem). For example, in even
an extreme address range of $L_{max}=10^{6}$ (in practice we rarely
need that big range) to train input sequences of length 10, we only
need to sample $10^{5}\log10^{5}$ sequences, which is often smaller
than the size of the training datasets. Empirically, in our experiments,
we always train our method with the same number of batch size and
training steps as other baselines to ensure a fair comparison, and
we realize that it is always possible to expose all the addresses
to our model during training.

\begin{table*}
\begin{centering}
{\small{}}%
\begin{tabular}{ccc}
\hline 
{\footnotesize{}Difference} & {\footnotesize{}Transformer} & {\footnotesize{}PANM (Our)}\tabularnewline
\hline 
{\footnotesize{}Key} & {\footnotesize{}Keys are computed based on input} & {\footnotesize{}The keys in our approach are }\tabularnewline
{\footnotesize{}Generation} & {\footnotesize{}data. Hence, when meeting novel } & {\footnotesize{}generated as fixed numbers, }\tabularnewline
 & {\footnotesize{}data during testing, Transformer} & {\footnotesize{}specifically physical memory}\tabularnewline
 & {\footnotesize{}will observe novel keys, and cannot} & {\footnotesize{}addresses. These keys are entirely}\tabularnewline
 & {\footnotesize{}work properly.} & {\footnotesize{}separate from the data.}\tabularnewline
\hline 
{\footnotesize{}Extendable} & {\footnotesize{}The dimension of attention weights } & {\footnotesize{}The fixed nature of our }\tabularnewline
{\footnotesize{}to Longer} & {\footnotesize{}varies with input length, making } & {\footnotesize{}physical addresses allows}\tabularnewline
{\footnotesize{}Sequences} & {\footnotesize{}arithmetic transformations on} & {\footnotesize{}our pointers to be easily }\tabularnewline
 & {\footnotesize{}these attention weights infeasible } & {\footnotesize{}manipulated and extendable}\tabularnewline
 & {\footnotesize{}as the sequence length increases.} & {\footnotesize{}to longer sequences.}\tabularnewline
\hline 
{\footnotesize{}Symbol} & {\footnotesize{}The use of attention weights } & {\footnotesize{}Using physical addresses as keys }\tabularnewline
{\footnotesize{}Processing} & {\footnotesize{}as implicit pointers may lack } & {\footnotesize{}in our approach is crucial for symbol}\tabularnewline
{\footnotesize{}Advantages} & {\footnotesize{}the explicitness needed for} & {\footnotesize{}processing as it explicitly allows}\tabularnewline
 & {\footnotesize{}effective symbol processing.} & {\footnotesize{}pointer assignment, dereference, }\tabularnewline
 &  & {\footnotesize{}and arithmetic operations.}\tabularnewline
\hline 
{\footnotesize{}Physical Address} & {\footnotesize{}Positional encoding can be generated} & {\footnotesize{}Our physical addresses are }\tabularnewline
{\footnotesize{}vs Positional} & {\footnotesize{}independently from data. However, } & {\footnotesize{}detached from the data, supporting}\tabularnewline
{\footnotesize{}Encoding} & {\footnotesize{}they are not separated from the input} & {\footnotesize{}the transformation of pointers}\tabularnewline
 & {\footnotesize{}data as our physical addresses. There} & {\footnotesize{}through timesteps and isolating}\tabularnewline
 & {\footnotesize{}is no explicit mechanism in} & {\footnotesize{}pointer manipulation from }\tabularnewline
 & {\footnotesize{}Transformer to attend only to these } & {\footnotesize{}the input.}\tabularnewline
 & {\footnotesize{}positional encodings or to transform } & \tabularnewline
 & {\footnotesize{}pointers to point to these positional } & \tabularnewline
 & {\footnotesize{}encodings from one step to another.} & \tabularnewline
\hline 
\end{tabular}{\small\par}
\par\end{centering}
\caption{PANM vs Transformer \label{tab:compare_T}}
\end{table*}

\subsection{More Discussion on Model and Architecture\label{subsec:Choice-of-Architecture}}

We can see that it is critical to have $H^{a}\geq2$ and $p_{B},p_{E}\in\left\{ p_{0,h}^{a}\right\} _{h=1}^{H^{a}}$
to achieve generalization using pointer arithmetic. In other words,
if $p_{B},p_{E}\notin\left\{ p_{0,h}^{a}\right\} _{h=1}^{H^{a}}$,
we can always find a task to make PANM fail to generalize. For example,
if $p_{E}\notin\left\{ p_{0,h}^{a}\right\} _{h=1}^{H^{a}}$, PANM
cannot generalize in Reverse task. To see that, without loss of generality,
we assume PANM only learn to produce the last token at address $p_{E}$
using information from some initial addresses $p'\in\left\{ p_{0,h}^{a}\right\} _{h=1}^{H^{a}}$
such that $p'\neq p_{E}$. During training, the learned pointer arithmetic
to perform Reverse at the first step of the decoding to produce value
$y_{1}$ can only be a function of $p'$: $y_{1}={}^{*}p_{E}=^{*}f(p')$,
that is, $p_{E}=f(p')$. During testing, $p_{E}$ can receive arbitrary
value, so for whatever learned $f$, we can always find a test sequence
such that $p_{E}\neq f(p')\,\,\forall f$ because $p_{E}\neq p'$.
A similar argument can be used for $p_{B}$ and Copy task. 

\begin{figure*}
\begin{centering}
\includegraphics[width=0.9\textwidth]{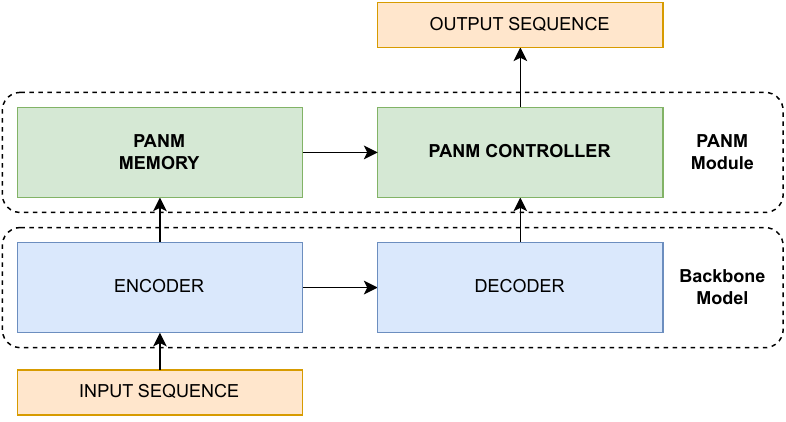}
\par\end{centering}
\caption{PANM as a plug-and-play architecture. The encoder and decoder can
be any model (LSTM, Transformer or BERT). PANM Controller can be used
as the last layer of the Decoder to access the memory during decoding.
To reduce the number of parameters of the augmented architecture,
the decoder's number of layers can be decreased. \label{fig:Model-architecture.-1}}
\end{figure*}

In the main manuscript, we only experiment with 1 Mode-2 pointer ($H^{c}=1$).
If $H^{c}=0$, obviously PANM will fail in tasks such as ID Sort.
Using more $H^{c}$ and $H^{a}$ can still be beneficial in exchange
for slower computation (see Appendix \ref{subsec:More-Experiments-with}).
In all experiments, we use 256-dimensional GRUs for the $\mathtt{PU}$
and $\mathtt{Ctrl}$. The encoder and decoder (to stack the Controller
on) can vary across tasks. The general plug-and-play framework is
illustrated in Fig. \ref{fig:Model-architecture.-1}. We also summarize
operations of our model in Algo. \ref{algo}.

\subsection{Experimental Details\label{subsec:Experimental-Details}}

All the datasets and public codebases use Apache or MIT License. We
trained all the models using a single GPU Tesla V100-SXM2. The running
time of PANM depends on the $\mathtt{Encoder}$ and tasks. Overall,
with 2 Mode-1 pointers and 1 Mode-2 pointer, PANM's speed will be
70-80\% compared to the backbone model. For example, in Copy task,
PANM's speed is 15 iterations/s while LSTM's is 20 iterations/s. If
PANM uses Transformer Encoder, its speed is 77 iterations/s while
Transformer's is 90 iterations/s.

\textbf{Baseline Choice }Although our baselines are classic, they
are still very strong baselines in our studied tasks. For example,
in our algorithmic reasoning, LSTM with attention or Pointer Networks
are still dominant baselines, outperforming the more recent Transformer.
In Dyck recognition, stack-based models are still SOTA because their
inductive bias is suitable for the task. Experiments in Sec. 3 adopt
(Universal) Transformer+RPE, which is a recent and strong Transformer
variant focusing on generalization. There are also other sophisticated
methods focusing generalization \cite{webb2020emergent}. 

In our experiments, PANM is ensured to have similar model size as
the baselines and often built on top of similar backbones for fair
comparison. We believe it is still important to improve fundamental
baselines such as Transformers or LSTM because they are the building
blocks of many practical applications including recent Large Language
Models (LLMs). In this paper, we prove the improvement of these fundamental
blocks, and in future works, we will extend our ideas to more advanced
backbones such as LLMs.

\subsubsection{Algorithmic Reasoning\label{subsec:Algorithmic-Reasoning}}

We first give the details of the content-based tasks below. 

In \textbf{Dynamic Recall}, an arbitrary input token is chosen as
the query and is added to the end of the input sequence. Depending
on the length of the input, a.k.a, odd or even, the first target token
will be on the left or right of the query, following its succeeding
tokens in the input. This task requires both content matching (find
query token in the input sequence) and position-based access (shift
left or right). 

In \textbf{Priority Sort}, each input token is associated with a priority
score sampled from the standard normal distribution. The target output
will be tokens from the input sequence sorted ascending by their the
score. This task can be solved in many ways and likely needs complicated
symbol processing such as looping through items in the sequence and
comparing the score of tokens. 

Finally, in \textbf{ID Sort}, each input token is augmented with an
id feature vector sampled from standard multivariate normal distribution
such that every 2 tokens share one id. For example, with input $x_{1},x_{2},x_{3},x_{4}$,
$x_{1}$ and $x_{4}$ may share one id while $x_{2}$ and $x_{3}$
shares another id. The pairing is chosen randomly. The output token
at position $i$-th will be the input token that share id with the
$i$-th input token. The correct output for the earlier example is
$x_{4},x_{3},x_{2},x_{1}$. This task is specifically designed to
test the ability to learn Mode 2 pointer-based memory access.

In this task, we implement the baselines such as LSTM, attention models
and Transformer using Pytorch library. The hidden state dimension
for these models are set to 512, which results in around 1-3 millions
parameters. We tuned the number of layers of the encoder/decoder for
these baselines in Copy task, and realized that 1-layer gave the best
performance. For NTM and DNC, we use public repositories\footnote{\url{https://github.com/thaihungle/SAM}}
with default controller's hidden state of 256 dimensions and 128-slot
external memory, which results in around 1.2 millions parameters.
We use the ESBN's author codebase \footnote{\url{https://github.com/taylorwwebb/emergent_symbols}}
with default parameter setting, resulting in $\approx$1.2 million
parameters. For PtrNet, since we do not use token index as the training
label, we produce the predicted token by performing weighted sum the
input tokens using the PtrNet's attention weights. PtrNet's hyperparameters
are the same as attention models. We could not find the authors' code
for Neural Stack and NRAM so we implemented them and tuned hyperparameters
for the Copy task at length $L$ such that the model sizes are about
1.1 million parameters. In this task PANM uses LSTM with hidden state
of 256 as the $\mathtt{Encoder}$ and does not stack the Controller
on any decoder models, resulting in $\approx$1.1 million parameters.

\begin{algorithm}[H]     
\SetAlgoLined     
\SetKwInput{KwInput}{Input} 
\SetKwInput{KwOutput}{Ouput} 
\SetKwComment{Comment}{/* }{ */}
\KwInput{A dataset of sequence pairs $D=\{X_i,Y_i\}_{i=1}^{N_{data}}$, initial $\Phi$ containing the parameters of the $\mathtt{Encoder}_\theta$, Pointer Unit $\mathtt{PU}_\varphi$ and Controller $\mathtt{Ctrl}_\lambda$,  $b$ representing the number of bits of the address space, $L_{dec}$ being the maximum number of decoding steps, and function $l$ measuing the length of a sequence.}
\KwOutput{$\Phi^*$, trained parameters.}
\For{$\{X_i, Y_i\}\sim D$}{
	\tcc{Construct the memory}
	{$\mathtt{M}=\mathtt{Encoder}_\theta(X_i)$}

	\tcc{Sample base address. During testing, $p_B$ can be set to 0}
	{$p_B \sim \mathrm{Uniform}\left(\{0,1\}^b\right)$}
	
	\tcc{Generate the address}
	\For{$j=0,1,...,l(\mathtt{M})-1$}{ 
		$\mathtt{A}[j]=(p_B+j)\ \mathtt{mod}\  2^b$ 
	}
	\tcc{Decode with pointers}
	\For{$t=0,1,...,L_{dec}$}{ 
		{Use $\mathtt{PU}_\varphi$ and $\mathtt{A}$ to compute $p^a_t$ using Eq. \ref{eq:pat}}

		{Use $\mathtt{M}$ and $p^a_t$ to compute pointer values $^{*}p_{t}^{a}$ (Mode 1) and $^{*}p_{t}^{c}$ (Mode 2) (see $\mathsection$\ref{subsec:Pointer-based-Addressing-Modes})}

		{Use $\mathtt{Ctr}_\lambda$ and pointer values to compute $p_{\Phi}(y_{t}|X_{i},z_{t})$ (see $\mathsection$\ref{subsec:The-Controller})}

		{$\hat{y}_{t}^{i}=\mathrm{argmax}_{y_t}\ p_{\Phi}(y_{t}|X_{i},z_{t})$}

		\If{$\hat{y}_{t}^{i}$ is EOS}{
		break
		}
	}
	\tcc{Compute cross-entropy loss}
	{$\mathcal{L}=-\sum_{t}\log p_{\Phi}(y_{t}=Y_i[t]|X_{i},y_{t-}^{i})$}

	{Use $\mathcal{L}$ to update $\Phi$ through backpropagation}
}
\caption{PANM training. To simplify, we assume the batch size and number of pointer heads of one.\label{algo}} 
\end{algorithm}

In this experiment, all the models are trained without teacher forcing
as in \citet{graves2014neural}, i.e, the input to the decoder is
zero ($z_{t}=0$). The detailed average accuracy (mean $\pm$ std.)
of each method together with the actual length of each testing mode
are reported in Tables \ref{tab:Copy:-accuracy-(mean}-\ref{tab:ID-Sort:-accuracy}. 

Overall, PANM observes significant improvement ranging from 10-20\%
on each task. We note that when compared with individual baselines,
the improvement is much higher. Consider Copy as an example (Fig.
\ref{fig:Examplar-performance-on}a), PANM outperforms the worst baseline
Transformer by around 60\% at $2(L+1)$ and 30\% at $4(L+1)$, respectively.
As stated earlier that our tasks are challenging, thus, originally
strong baselines such as NTM, DNC, and Neural Stack do not generalize
well at extreme lengths, especially in ID Sort. ID Sort is trickier
than content-free tasks, making some baselines fail at length $L$
even though it is in the training data. The best other model in this
case is Content Attention, which clearly underperforms our PANM from
few \% to 50\% (Fig. \ref{fig:Examplar-performance-on}b). Without
curriculum learning and under the 10-class prediction setting, methods
that use implicit pointers, such as PtrNet, NRAM, and ESBN, demonstrate
mediocre performance on average when compared to PANM. Furthermore,
PANM also outperforms in length-dependent tasks (Mix, D. Recall),
indicating that it can track the sequence length in extrapolation.
We hypothesize that PANM's content--free pointer generation mechanism
to simulate list iteration makes it possible. 

In Copy, only Mode-1 access is needed. As decoding step $t$ increases,
Pointer Unit generates $p_{t}^{a}$ following the increment of the
addresses as expected. That said, for several steps, the address attention
is not sharp, showing other addresses pointed by the pointer, which
is not surprising since we use soft attention and it is hard for a
neural network to learn the exact rule: $p_{t+1}^{a}=p_{t}^{a}+1$.
This problem gets worse as test length increases as the error accumulates,
especially when the same token can appear many times, which confuses
the model. This explains why PANM's performance drops clearly in the
hardest case $8(L+1$). Yet, it is still significantly better than
others whose results are near random prediction. 

\subsubsection{Dyck Language Recognition\label{subsec:Dyck-Language-Recognition-1}}

\begin{figure*}
\begin{centering}
\includegraphics[width=0.9\linewidth]{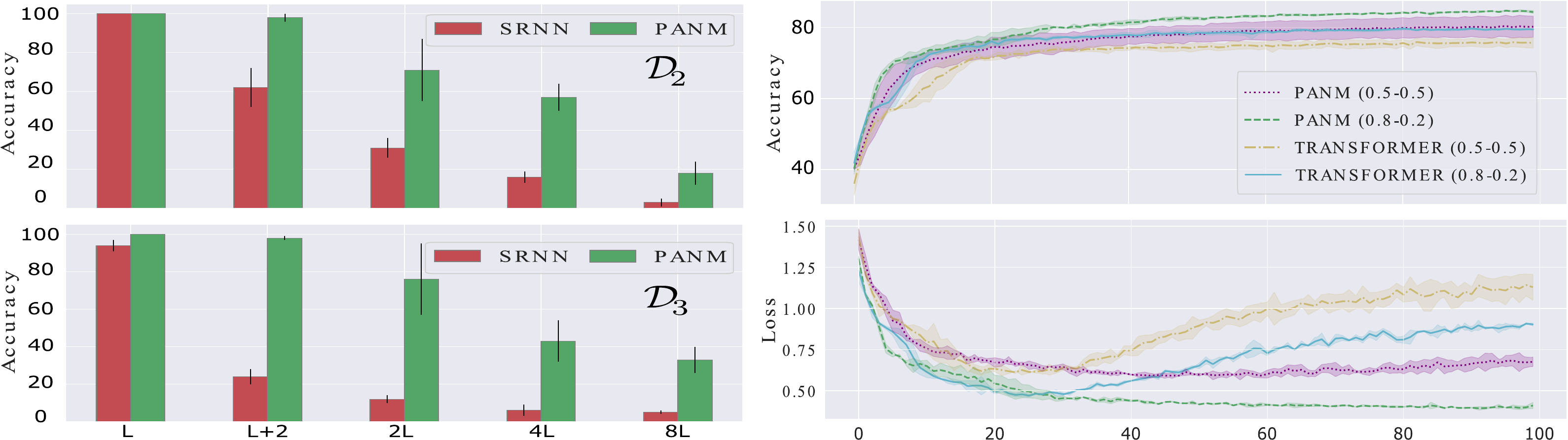}
\par\end{centering}
\caption{Dyck (Left): mean $\pm$ std. accuracy over 5 runs with different
testing lengths. bAbI QA (Right): mean $\pm$ std. testing accuracy
and cross-entropy loss across 100 training epochs over 5 runs. \label{fig:dyck}\label{fig:Testing-curves-of}}
\end{figure*}

In this task, we adopt the SRNN code from \citet{suzgun2019memory}\footnote{\url{https://github.com/suzgunmirac/marnns}}
using the default parameters. As explained in the main text, this
task is auto-regression, hence, $z_{t}=\hat{y}_{t-1}$. PANM adopts
SRNN (an auto-regressive model) as the encoder and does not stack
the Controller on any decoder models. The result is visualized in
Fig. \ref{fig:dyck} (left).

\subsubsection{Conpositional Learning\label{subsec:Conpositional-Learning-App}}

In this task, we adopt the code from \citet{csordas2021devil}\footnote{\url{https://github.com/RobertCsordas/transformer_generalization}}
using the default parameters. When using Transformer Encoder, we need
to have Transformer-like decoder to align the token representation
of the encoding and decoding phases. As such, in SCAN, we utilize
the 3-layer Transformer decoder, replace its last layer by the Controller.
Formally, $z_{t}$ in Eq. \ref{eq:ctr} becomes Decoder($y_{t-}$)
where the Decoder is a 2-layer Transformer. In Mathematics reasoning
task, we use similar integration except that the Encoder is Transformer
with relative positional encoding (TRM + RPE). By reducing the number
of decoding layer, we ensure PANM's hyperparameter number equivalent
to that of the Transformer baseline (12M). All models are trained
with teacher forcing as in \citet{csordas2021devil}. 

\begin{figure*}
\begin{centering}
\includegraphics[width=0.95\linewidth]{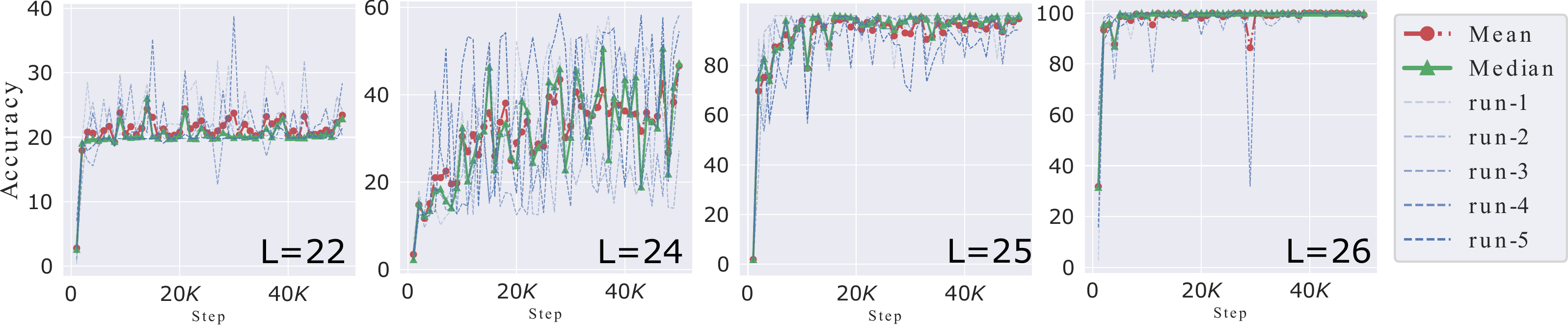}
\par\end{centering}
\caption{SCAN: PANM's exemplar learning curves.\label{fig:SCAN:-PANM's-exemplar} }
\end{figure*}

\textbf{SCAN} The training size is 16990 and the test size is 3920.
SCAN is a well-known and standard benchmark for testing compositional
learning and generalization in sequential models. One property of
this dataset is that a new length often contains new rules that must
be captured by the model to ensure generalization, and thus, if the
model fails to learn a hidden rule, its performance may drop significantly
from one length split to another. Fig. \ref{fig:SCAN:-PANM's-exemplar}
illustrates PANM's testing accuracy curves when $L=22,24,25,26$.
Other learning curves for $L>26$ looks similar to $L=26$ where PANM
easily solves the task perfectly. 

\textbf{Mathematical Problems} Table \ref{tab:Mathematics-dataset:-mean-1}
reports the accuracy with mean and standard deviation. Here, we augment
TRM and TRM+RPE with PANM. Both shows improvement, especially for
TRM+RPE, indicating that PANM is compatible with other methods designed
to improve generalization in Transformer. 

\subsubsection{Other NLP Tasks \label{subsec:Question-Answering-App}}

The bAbI dataset consists of 20 synthetic tasks that evaluate various
reasoning skills. To prepare the data for each task, we combine train/valid/test
into a single set and sort it by length and split it into training
and testing sets, as described in the main text. We train the models
jointly on all 20 tasks and measure the accuracy of their answers,
which are considered correct only if they match the ground truth answer
perfectly. The training/evaluation follows exactly the standard protocol
presented in \cite{pmlr-v119-le20b}. The Transformer used here has
8 heads, 3 layers of encoding, 3 layers of decoding, and hidden dimensions
of 512. PANM uses the same Transformer backbone except that the decoder
has 2 layers to make the model size equivalent. We run each model
5 times to report the mean and standard deviation as in Fig. \ref{fig:Testing-curves-of}
(right). Table \ref{tab:bAbI-QA:-testing} reports the detailed numbers. 

The SQUAD dataset contains more than 100K realistic context/question-answer
pairs. Again, we combine train/test into a single set and sort it
by length and split into new train/test sets. Following \citet{kenton2019bert},
we use BERT model (\url{https://huggingface.co/bert-base-uncased})
to predict the start and end location of the answer, and finetune
the model with the same setting (e.g., 3 epochs with a learning rate
of 5e-5) except that our batch size is 16 to fit with our GPU. PANM
appends the Controller to BERT to predict the start and end. Both
BERT and PANM have around 114 million parameters. Table \ref{tab:bAbI-QA:-testing-1}
reports the detailed numbers. 

\subsubsection{Additional Experiments\label{subsec:More-Experiments-with}}

\begin{figure}
\begin{centering}
\includegraphics[width=0.95\linewidth]{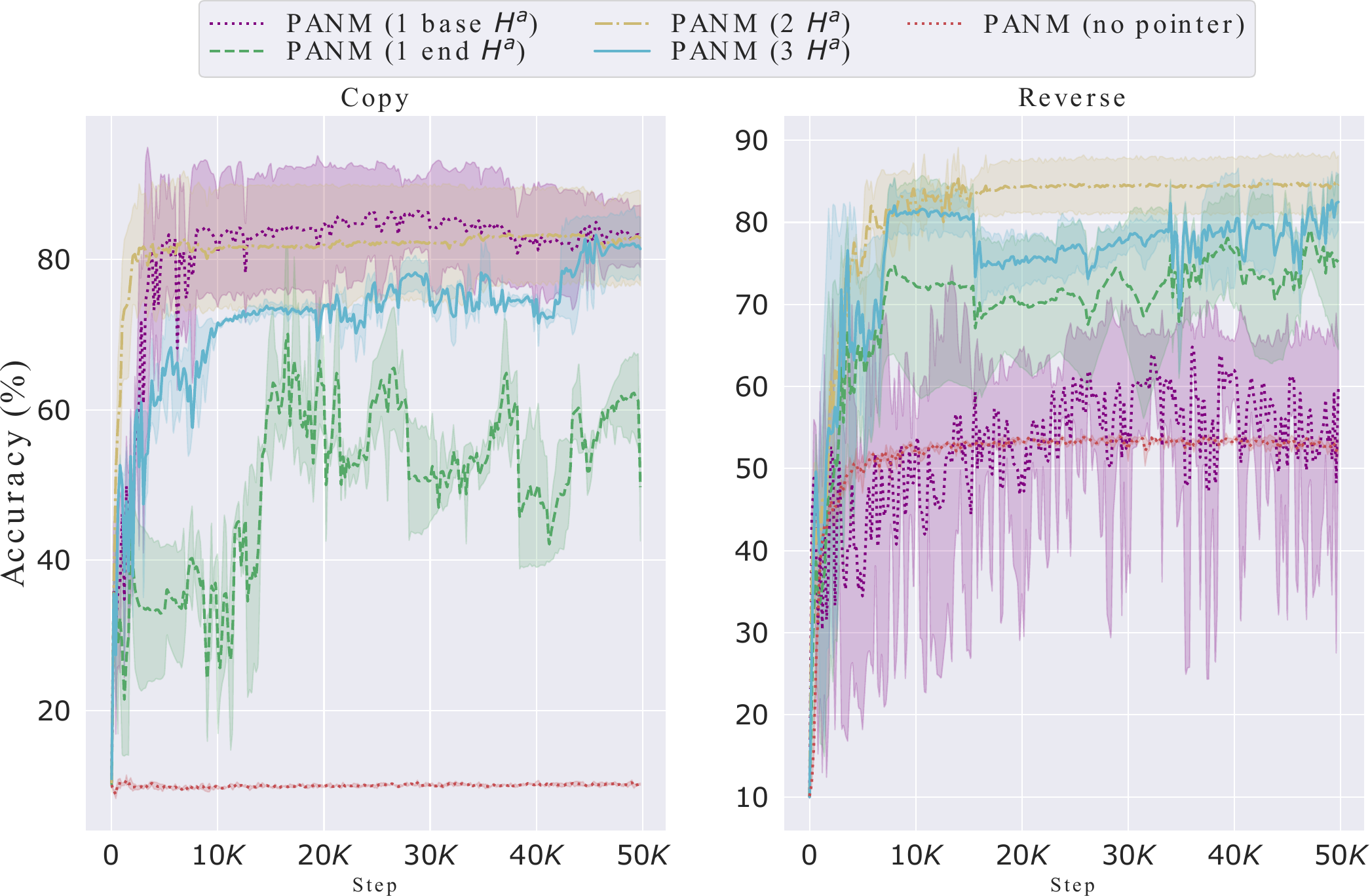}
\par\end{centering}
\caption{Testing accuracy (mean $\pm$ std.) at 2(L+1) length over training
steps. Different configurations of Mode-1 pointers are trained and
evaluated 5 times.\label{fig:Testing-accuracy-at} }
\end{figure}
\begin{figure*}
\begin{centering}
\includegraphics[width=0.95\linewidth]{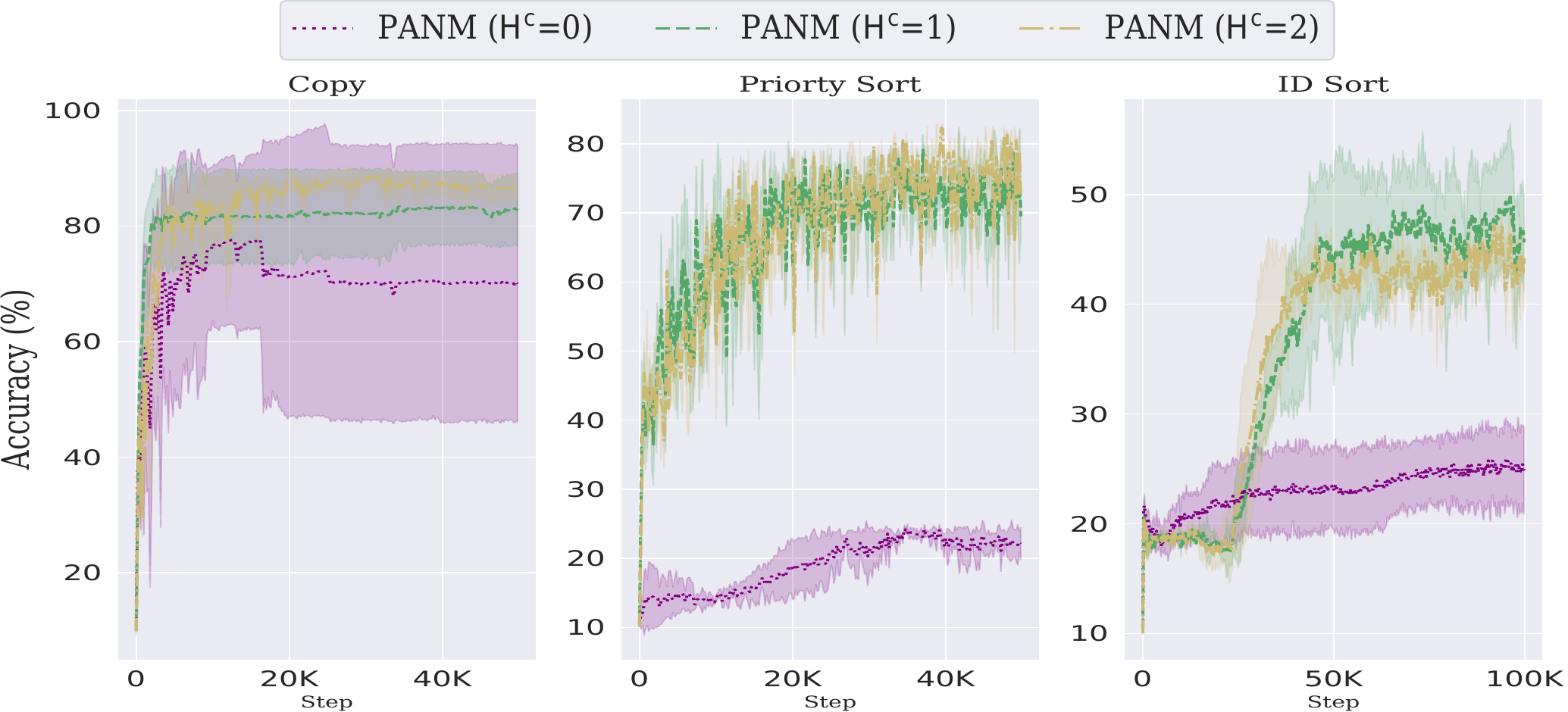}
\par\end{centering}
\caption{Testing accuracy (mean $\pm$ std.) at 2(L+1) length over training
steps. Different configurations of Mode-2 pointers are trained and
evaluated 5 times.\label{fig:Testing-accuracy-at-1}}
\end{figure*}

\paragraph{Pointer Hyperparameters}

In this section, we confirm the logic presented in Appendix \ref{subsec:Choice-of-Architecture}
by performing experiments that involve varying the number and type
of pointers.\textbf{ }

\textbf{Mode-1 Pointers }We test the PANM version in $\mathsection$
\ref{subsec:Algorithmic-Reasoning-1} with $H^{a}=0,1,2,3$ on Copy,
Reverse. We do not use Mode-2 pointer here to avoid confusion ($H^{c}=0$).
Fig. \ref{fig:Testing-accuracy-at} plots the testing accuracy over
training time. As $H^{a}=0$, there is no pointer information for
the Controller, PANM should be equivalent to an GRU and fail to generalize.
As $H^{a}=1$, the only pointer is initialized either with the base
or end address. As shown in Fig., PANM cannot generalize in both Copy
and Reverse tasks with single Mode-1 pointer, which is proved in Appendix
\ref{subsec:Choice-of-Architecture}. In the case$H^{a}=3$, we initialize
them with the base, end and middle addresses. We observe that increasing
$H^{a}$ to 3 slightly reduces the performance in these tasks. We
speculate that too many Mode-1 pointers make the learning harder;
in particular, learning to manipulate the third pointer may interfere
with that of the first or second pointer, which are more important
in these tasks. Generally, most tasks only require list iterations
from the head to the tail or vice versa. Hence, we keep $H^{a}=2$
in all experiments to save the computation cost. 

\textbf{Mode-2 Pointers }We fix $H^{a}=2$, and vary $H^{c}=0,1,2$
on Copy, Priority Sort, ID Sort. As shown in Fig. \ref{fig:Testing-accuracy-at-1},
without Mode-2 pointers ($H^{c}=0$), generalization in Priority Sort
and ID Sort is reduced significantly by 50\% and 30\%, respectively
because these tasks focus more on the content of the input sequence
and often demand comparing the content of different tokens. Interestingly,
a content-free task like Copy also suffers from performance drop if
there is no Mode-2 pointer. Specifically, we find out that for 2/5
runs, the model converges to a suboptimal solution, leading to high
variance and slightly lower mean accuracy. Perhaps, Mode-2 pointer
allows the decoder to access the input instantly (like content-based
attention), avoid forgetting, and thus, improve the prediction as
the sequence is longer. Having more Mode-2 pointers generally improves
the generalization in Copy and Priority Sort, yet the gain is small
for $H^{c}=2$, or even negative in ID Sort. Therefore, we trade-off
performance with computing efficiency by setting $H^{c}=1$ in our
experiments. 

\subsubsection{Failures of Chat-GPT in Our Tasks\label{subsec:Failures-of-Chat-GPT}}

Large Language Models (LLMs), especially Chat-GPT, have shown remarkable
results in reasoning and generalization. Directly comparing Chat-GPT
with other models used in our experiments would be unfair because
Chat-GPT was not directly trained with our datasets and it has much
more parameters than our model. Therefore, in this section, we merely
use Chat-GPT as a tool to verify that our chosen tasks, despite being
simple, are non-trivial. The evaluated tasks are algorithmic reasoning
and SCAN. We do not examine Dyck recognition because the output encoding
is complicated to represent in text. Other datasets are more common
and likely to be used for training Chat-GPT, thus, are not suitable
for generalization test. For example, in Mathematics task, if we ask
Chat-GPT the question from the data $\mathtt{What\,\,is\,\,the\,\,hundreds\,\,digit\,\,of\,\,31253?}$,
it provide the correct answer ($\mathtt{2}$). However, slightly modifying
the question to ensure it does not appear in the training and testing
set will successfully fool Chat-GPT:
\begin{itemize}
\item Example 1:
\begin{itemize}
\item Prompt: What is the hundreds digit of 312537?
\item Chat-GPT answer: The hundreds digit of the number 312537 is 2.
\end{itemize}
\item Example 2:
\begin{itemize}
\item Prompt: What is the hundreds digit of 319253?
\item Chat-GPT answer: The hundreds digit of the number 319253 is 9.
\end{itemize}
\end{itemize}
We use Open AI's Chat-GPT 3.5 version September and evaluate the model
on our data using few-shot example prompts, following the format:

Examples: \\
input $x^1_1,x^1_2,...$ output $y^1_1,y^1_2,...$ \\
input  $x^2_1,x^2_2,...$ output $y^2_1,y^2_2,...$ \\
... \\
Question: \\
input $x_1,x_2,...$ output 

\paragraph{Algorithmic Reasoning }

To ensure that Chat-GPT does not memorize the output answer from its
vast training data, we use non-digit symbols: \textasciitilde !@\#\$\%\textasciicircum\&{*}(
as 10 tokens of the datasets. For each task, we sample 20 training
examples of length $L=5$ to build the in-context examples, and test
on 1 longer sequence of length $2L=10$. We conduct 20 trials and
report the average test accuracy. Table \ref{tab:Failure-of-Chat-GPT}
summaries the evaluation results. Overall, except for Copy task where
Chat-GPT shows excellent generalization, other tasks are very hard
for Chat-GPT, indicating that the length extrapolation problem still
poses a big challenge to today AI techniques. 

\begin{table*}
\begin{centering}
{\small{}}%
\begin{tabular}{cccccc}
\hline 
\multirow{2}{*}{{\small{}Task}} & \multirow{2}{*}{{\small{}Chat-GPT }} & \multicolumn{3}{c}{{\small{}Failure example}} & \multirow{2}{*}{{\small{}PANM }}\tabularnewline
 &  & {\small{}Input} & {\small{}Chat-GPT Output} & {\small{}True Output} & \tabularnewline
\hline 
{\small{}Copy} & {\small{}100\%} & \multicolumn{3}{c}{{\small{}N/A}} & {\small{}84\%}\tabularnewline
{\small{}Reverse} & {\small{}69\%} & {\small{}\$\%\&\&\$\%\textasciicircum @\%\#} & {\small{}\%\#\textasciicircum @\textasciicircum\%\$\&\&\%\$} & {\small{}\#\%@\textasciicircum\%\$\&\&\%\$} & {\small{}84\%}\tabularnewline
{\small{}Mix} & {\small{}42\%} & {\small{}\$\%\&\&\$\%\textasciicircum @\%\#} & {\small{}\%\#\textasciicircum\&\&\$\%\$@\%\&} & {\small{}\$\%\$\&\$\%\$@\$\#} & {\small{}98\%}\tabularnewline
{\small{}Dynamic Recall} & {\small{}14\%} & {\small{}\$(\&\&\$\#\textasciicircum @\%\# \%} & {\small{}\$} & {\small{}@} & {\small{}45\%}\tabularnewline
\hline 
\end{tabular}{\small\par}
\par\end{centering}
\caption{Failure of Chat-GPT on algorithmic reasoning test cases of length
$2L$. Token-level accuracy is reported. We do not test Chat-GPT on
Priority and ID sort because they have complicated token representations.
PANM results cannot be directly compared, and shown for reference
only. \label{tab:Failure-of-Chat-GPT}}

\end{table*}

\paragraph{SCAN}

In this task, we sample 20 examples in the L-cutoff=40 split set (easiest)
as in-context learning examples and evaluate on 10 unseen sequences.
Chat-GPT totally failed in this task. When testing on the similar
length or longer length as the examples, Chat-GPT cannot produce any
exact match results (exact match accuracy=0). Below are some failure
examples:
\begin{itemize}
\item IN: walk and turn opposite right OUT: 
\begin{itemize}
\item Chat-GPT output: I\_TURN\_RIGHT I\_TURN\_RIGHT I\_WALK
\item True output: I\_WALK I\_TURN\_RIGHT I\_TURN\_RIGHT
\end{itemize}
\item IN: run around left twice and run around right OUT: 
\begin{itemize}
\item Chat-GPT output: I\_RUN I\_TURN\_LEFT I\_RUN I\_TURN\_LEFT I\_RUN
I\_TURN\_RIGHT I\_RUN
\item True output: I\_TURN\_LEFT I\_RUN I\_TURN\_LEFT I\_RUN I\_TURN\_LEFT
I\_RUN I\_TURN\_LEFT I\_RUN I\_TURN\_LEFT I\_RUN I\_TURN\_LEFT I\_RUN
I\_TURN\_LEFT I\_RUN I\_TURN\_LEFT I\_RUN I\_TURN\_RIGHT I\_RUN I\_TURN\_RIGHT
I\_RUN I\_TURN\_RIGHT I\_RUN I\_TURN\_RIGHT I\_RUN
\end{itemize}
\end{itemize}

\paragraph{}

\begin{table*}
\begin{centering}
\begin{tabular}{cccccc}
\hline 
Copy Task & 9(L) & 10(L+1) & 20((L+1){*}2) & 40((L+1){*}4) & 80((L+1){*}8)\tabularnewline
\hline 
LSTM & 100$\pm$0 & 47$\pm$0 & 11$\pm$0 & 10$\pm$0 & 10$\pm$0\tabularnewline
Location Attention & 100$\pm$0 & 93$\pm$2 & 51$\pm$5 & 28$\pm$4 & 20$\pm$1\tabularnewline
Content Attention & 100$\pm$0 & 92$\pm$1 & 53$\pm$0 & 33$\pm$0 & 22$\pm$0\tabularnewline
Hybrid Attention & 100$\pm$0 & 91$\pm$1 & 50$\pm$1 & 23$\pm$3 & 13$\pm$0\tabularnewline
Transformer & 100$\pm$0 & 20$\pm$1 & 16$\pm$0 & 13$\pm$0 & 11$\pm$0\tabularnewline
NTM & 100$\pm$0 & 74$\pm$4 & 13$\pm$2 & 11$\pm$0 & 11$\pm$0\tabularnewline
DNC & 100$\pm$0 & 54$\pm$2 & 11$\pm$1 & 11$\pm$0 & 11$\pm$0\tabularnewline
Neural Stack & 100$\pm$0 & 90$\pm$4 & 47$\pm$2 & 29$\pm$0 & 17$\pm$0\tabularnewline
PtrNet & 100$\pm$0 & 90$\pm$2 & 52$\pm$1 & 32$\pm$1 & 20$\pm$0\tabularnewline
NRAM & 100$\pm$0 & 81$\pm$3 & 15$\pm$2 & 11$\pm$0 & 11$\pm$1\tabularnewline
ESBN & 100$\pm$0 & 92$\pm$0 & 34$\pm$0 & 11$\pm$0 & 11$\pm$0\tabularnewline
\hline 
PANM & \textbf{100$\pm$0} & \textbf{100$\pm$0} & \textbf{84$\pm$1} & \textbf{52$\pm$1} & \textbf{36$\pm$1}\tabularnewline
\hline 
\end{tabular}
\par\end{centering}
\caption{Copy: accuracy (mean $\pm$ std. over 5 runs)\label{tab:Copy:-accuracy-(mean}}
\end{table*}
\begin{table*}
\begin{centering}
\begin{tabular}{cccccc}
\hline 
Reverse Task & 9(L) & 10(L+1) & 20((L+1){*}2) & 40((L+1){*}4) & 80((L+1){*}8)\tabularnewline
\hline 
LSTM & 100$\pm$0 & 96$\pm$0 & 53$\pm$0 & 33$\pm$0 & 22$\pm$0\tabularnewline
Location Attention & 100$\pm$0 & 26$\pm$3 & 18$\pm$1 & 14$\pm$0 & 12$\pm$0\tabularnewline
Content Attention & 100$\pm$0 & 81$\pm$25 & 38$\pm$11 & 23$\pm$4 & 16$\pm$2\tabularnewline
Hybrid Attention & 100$\pm$0 & 98$\pm$1 & 50$\pm$7 & 24$\pm$2 & 15$\pm$1\tabularnewline
Transformer & 100$\pm$0 & 18$\pm$0 & 15$\pm$3 & 13$\pm$1 & 11$\pm$0\tabularnewline
NTM & 100$\pm$0 & 95$\pm$7 & 65$\pm$27 & 26$\pm$13 & 13$\pm$1\tabularnewline
DNC & 100$\pm$0 & 93$\pm$3 & 60$\pm$18 & 23$\pm$6 & 12$\pm$1\tabularnewline
Neural Stack & 100$\pm$0 & 96$\pm$1 & 64$\pm$4 & 35$\pm$3 & 19$\pm$1\tabularnewline
PtrNet & 100$\pm$0 & 77$\pm$5 & 32$\pm$4 & 22$\pm$1 & 12$\pm$0\tabularnewline
NRAM & 100$\pm$0 & 96$\pm$1 & 60$\pm$3 & 33$\pm$2 & 15$\pm$2\tabularnewline
ESBN & 99$\pm$0 & 95$\pm$0 & 14$\pm$2 & 11$\pm$0 & 10$\pm$0\tabularnewline
\hline 
PANM & \textbf{100$\pm$0} & \textbf{100$\pm$0} & \textbf{84$\pm$3} & \textbf{51$\pm$1} & \textbf{33$\pm$1}\tabularnewline
\hline 
\end{tabular}
\par\end{centering}
\caption{Reverse: accuracy (mean $\pm$ std. over 5 runs)}
\end{table*}
\begin{table*}
\begin{centering}
\begin{tabular}{cccccc}
\hline 
Mix Task & 9(L) & 10(L+1) & 20((L+1){*}2) & 40((L+1){*}4) & 80((L+1){*}8)\tabularnewline
\hline 
LSTM & 100$\pm$0 & 96$\pm$0 & 53$\pm$0 & 33$\pm$0 & 22$\pm$0\tabularnewline
Location Attention & 100$\pm$0 & 92$\pm$10 & 56$\pm$1 & 45$\pm$0 & 30$\pm$6\tabularnewline
Content Attention & 100$\pm$0 & 61$\pm$8 & 57$\pm$1 & 14$\pm$0 & 12$\pm$0\tabularnewline
Hybrid Attention & 100$\pm$0 & 98$\pm$1 & 56$\pm$3 & 34$\pm$0 & 23$\pm$6\tabularnewline
Transformer & 100$\pm$0 & 18$\pm$0 & 15$\pm$3 & 13$\pm$1 & 11$\pm$0\tabularnewline
NTM & 100$\pm$0 & 95$\pm$7 & 65$\pm$27 & 26$\pm$13 & 13$\pm$1\tabularnewline
DNC & 100$\pm$0 & 91$\pm$4 & 58$\pm$9 & 19$\pm$3 & 11$\pm$1\tabularnewline
Neural Stack & 100$\pm$0 & 87$\pm$3 & 50$\pm$5 & 14$\pm$2 & 11$\pm$0\tabularnewline
PtrNet & 100$\pm$0 & 59$\pm$3 & 51$\pm$3 & 13$\pm$1 & 11$\pm$0\tabularnewline
NRAM & 99$\pm$0 & 82$\pm$7 & 48$\pm$6 & 17$\pm$4 & 10$\pm$1\tabularnewline
ESBN & 99$\pm$0 & 95$\pm$0 & 14$\pm$2 & 11$\pm$0 & 10$\pm$0\tabularnewline
\hline 
PANM & \textbf{100$\pm$0} & \textbf{100$\pm$0} & \textbf{98$\pm$1} & \textbf{54$\pm$0} & \textbf{54$\pm$1}\tabularnewline
\hline 
\end{tabular}
\par\end{centering}
\caption{Mix: accuracy (mean $\pm$ std. over 5 runs)}
\end{table*}
\begin{table*}
\begin{centering}
\begin{tabular}{cccccc}
\hline 
Drecall Task & 9(L) & 10(L+1) & 20((L+1){*}2) & 40((L+1){*}4) & 80((L+1){*}8)\tabularnewline
\hline 
LSTM & 85$\pm$7 & 74$\pm$16 & 21$\pm$2 & 12$\pm$1 & 11$\pm$0\tabularnewline
Location Attention & 88$\pm$1 & 82$\pm$1 & 30$\pm$3 & 19$\pm$2 & 13$\pm$0\tabularnewline
Content Attention & 88$\pm$2 & 84$\pm$0 & 27$\pm$3 & 17$\pm$1 & 13$\pm$1\tabularnewline
Hybrid Attention & 69$\pm$25 & 66$\pm$24 & 28$\pm$4 & 19$\pm$2 & 13$\pm$1\tabularnewline
Transformer & 33$\pm$1 & 32$\pm$0 & 22$\pm$0 & 14$\pm$1 & 12$\pm$1\tabularnewline
NTM & 86$\pm$3 & 72$\pm$8 & 22$\pm$1 & 15$\pm$0 & 12$\pm$0\tabularnewline
DNC & 89$\pm$0 & 83$\pm$1 & 22$\pm$1 & 14$\pm$2 & 11$\pm$0\tabularnewline
Neural Stack & 85$\pm$4 & 76$\pm$2 & 23$\pm$1 & 15$\pm$1 & 13$\pm$1\tabularnewline
PtrNet & 65$\pm$14 & 48$\pm$7 & 25$\pm$6 & 14$\pm$1 & 12$\pm$1\tabularnewline
NRAM & 61$\pm$6 & 59$\pm$4 & 21$\pm$4 & 13$\pm$2 & 11$\pm$1\tabularnewline
ESBN & 90$\pm$1 & 86$\pm$1 & 22$\pm$3 & 11$\pm$1 & 10$\pm$0\tabularnewline
\hline 
PANM & \textbf{92$\pm$0} & \textbf{89$\pm$0} & \textbf{45$\pm$1} & \textbf{22$\pm$0} & \textbf{16$\pm$0}\tabularnewline
\hline 
\end{tabular}
\par\end{centering}
\caption{Drecall: accuracy (mean $\pm$ std. over 5 runs)}
\end{table*}
\begin{table*}
\begin{centering}
\begin{tabular}{cccccc}
\hline 
PSort Task & 10(L) & 11(L+1) & 21((L+1){*}2) & 41((L+1){*}4) & 81((L+1){*}8)\tabularnewline
\hline 
LSTM  & 87$\pm$2 & 83$\pm$2 & 28$\pm$3 & 16$\pm$1 & 12$\pm$1\tabularnewline
Location Attention & 69$\pm$3 & 66$\pm$3 & 45$\pm$1 & 27$\pm$2 & 20$\pm$2\tabularnewline
Content Attention & \textbf{97$\pm$0} & 96$\pm$0 & 57$\pm$6 & 30$\pm$7 & 22$\pm$5\tabularnewline
Hybrid Attention & 85$\pm$3 & 81$\pm$1 & 33$\pm$1 & 25$\pm$2 & 23$\pm$3\tabularnewline
Transformer & 71$\pm$9 & 48$\pm$8 & 21$\pm$3 & 16$\pm$4 & 14$\pm$4\tabularnewline
NTM & 96$\pm$2 & 95$\pm$3 & 34$\pm$18 & 12$\pm$1 & 10$\pm$0\tabularnewline
DNC & 95$\pm$0 & 92$\pm$2 & 29$\pm$7 & 11$\pm$1 & 10$\pm$0\tabularnewline
Neural Stack & 92$\pm$2 & 79$\pm$2 & 32$\pm$3 & 13$\pm$2 & 11$\pm$1\tabularnewline
PtrNet & 77$\pm$2 & 71$\pm$2 & 43$\pm$2 & 24$\pm$1 & 19$\pm$1\tabularnewline
NRAM & 82$\pm$3 & 80$\pm$2 & 51$\pm$2 & 25$\pm$1 & 13$\pm$1\tabularnewline
ESBN & 26$\pm$4 & 24$\pm$4 & 13$\pm$2 & 11$\pm$1 & 10$\pm$0\tabularnewline
\hline 
PANM & \textbf{97$\pm$0} & \textbf{97$\pm$1} & \textbf{86$\pm$2} & \textbf{32$\pm$7} & \textbf{27$\pm$4}\tabularnewline
\hline 
\end{tabular}
\par\end{centering}
\caption{PSort: accuracy (mean $\pm$ std. over 5 runs)}
\end{table*}
\begin{table*}
\begin{centering}
\begin{tabular}{cccccc}
\hline 
ID Sort Task & 10(L) & 11(L+1) & 21((L+1){*}2) & 41((L+1){*}4) & 81((L+1){*}8)\tabularnewline
\hline 
LSTM  & 48$\pm$10 & 40$\pm$5 & 20$\pm$1 & 13$\pm$1 & 11$\pm$1\tabularnewline
Location Attention & 34$\pm$1 & 32$\pm$1 & 20$\pm$0 & 14$\pm$0 & 12$\pm$0\tabularnewline
Content Attention & 98$\pm$1 & 56$\pm$1 & 28$\pm$2 & 16$\pm$0 & 12$\pm$0\tabularnewline
Hybrid Attention & 32$\pm$1 & 31$\pm$1 & 19$\pm$1 & 14$\pm$0 & 12$\pm$0\tabularnewline
Transformer & 34$\pm$2 & 29$\pm$0 & 19$\pm$0 & 15$\pm$0 & 12$\pm$0\tabularnewline
NTM & 40$\pm$23 & 32$\pm$17 & 16$\pm$4 & 12$\pm$2 & 11$\pm$0\tabularnewline
DNC & 35$\pm$1 & 36$\pm$1 & 23$\pm$2 & 17$\pm$2 & 13$\pm$0\tabularnewline
Neural Stack & 33$\pm$3 & 32$\pm$1 & 19$\pm$1 & 13$\pm$1 & 12$\pm$0\tabularnewline
PtrNet & 27$\pm$1 & 24$\pm$1 & 15$\pm$0 & 12$\pm$1 & 11$\pm$0\tabularnewline
NRAM & 31$\pm$2 & 29$\pm$1 & 14$\pm$0 & 12$\pm$0 & 11$\pm$0\tabularnewline
ESBN & 47$\pm$18 & 42$\pm$12 & 18$\pm$0 & 12$\pm$0 & 10$\pm$0\tabularnewline
\hline 
PANM & \textbf{100$\pm$0} & \textbf{100$\pm$0} & \textbf{56$\pm$2} & \textbf{25$\pm$0} & \textbf{15$\pm$0}\tabularnewline
\hline 
\end{tabular}
\par\end{centering}
\caption{ID Sort: accuracy (mean $\pm$ std. over 5 runs)\label{tab:ID-Sort:-accuracy}}
\end{table*}
\begin{table*}
\begin{centering}
\begin{tabular}{ccc}
\hline 
Task & $\mathtt{add\_or\_sub}$ & $\mathtt{place\_value}$\tabularnewline
\hline 
U. TRM+ RPE$^{\clubsuit}$ & \textbf{0.97 \textpm{} 0.01} & 0.75 \textpm{} 0.10\tabularnewline
TRM + RPE$^{\clubsuit}$ & 0.91 \textpm{} 0.03 & -\tabularnewline
TRM + RPE$^{\diamondsuit}$ & 0.91 \textpm{} 0.04 & 0 \textpm{} 0\tabularnewline
TRM$^{\clubsuit}$ & 0.89 \textpm{} 0.01 & 0.12 \textpm{} 0.07\tabularnewline
TRM$^{\diamondsuit}$ & 0.86 \textpm{} 0.01 & 0.05+0.05\tabularnewline
U. TRM$^{\clubsuit}$ & 0.94 \textpm{} 0.01 & 0.20 \textpm{} 0.02\tabularnewline
\hline 
PANM TRM base (Ours) & 0.91 \textpm{} 0.01 & 0.15 \textpm{} 0.02\tabularnewline
PANM TRM + RPE base (Ours) & \textbf{0.97 \textpm{} 0.02} & \textbf{0.86 \textpm{} 0.05}\tabularnewline
\hline 
\end{tabular}
\par\end{centering}
\caption{Mathematics: mean \textpm{} std accuracy over 5 runs. $\clubsuit$
are numbers from \citet{csordas2021devil}. $\diamondsuit$ is our
rerun to confirm the results, which, in some cases, could not match
the reported numbers. - means training crash reported in the original
papers. We run PANM using the authors' codebase. \label{tab:Mathematics-dataset:-mean-1}}
\end{table*}

\end{document}